\documentclass{article}

\usepackage[final,nonatbib]{nips_2016_modded}

\usepackage[utf8]{inputenc} % allow utf-8 input
\usepackage[T1]{fontenc}    % use 8-bit T1 fonts
\usepackage{hyperref}       % hyperlinks
\usepackage{url}            % simple URL typesetting
\usepackage{booktabs}       % professional-quality tables
\usepackage{amsmath,amsfonts,amssymb}       % blackboard math symbols
\usepackage{nicefrac}       % compact symbols for 1/2, etc.
\usepackage{microtype}      % microtypography
\usepackage{bm}      % microtypography
\usepackage[pdftex]{graphicx}
\usepackage{algorithm}
\usepackage[noend]{algpseudocode}
\usepackage{tablefootnote}
\usepackage{pifont}

\makeatletter
\def\algbackskip{\hskip-\ALG@thistlm}
\makeatother

\usepackage{color}

\newcommand{\ignore}[1]{}

\renewcommand{\k}{\bm{k}}

\newcommand{\x}{\bm{x}}
\newcommand{\xone}{\bm{x}^{(1)}}
\newcommand{\xii}{\bm{x}^{(i)}}
\newcommand{\xj}{\bm{x}^{(j)}}
\newcommand{\xn}{\bm{x}^{(n)}}
\newcommand{\xsearch}{\x_{\text{search}}}

\newcommand{\yone}{y^{(1)}}
\newcommand{\yii}{y^{(i)}}

\newcommand{\yn}{y^{(n)}}

\newcommand{\zii}{z^{(i)}}

\newcommand{\xx}{\text{\textbf{X}}}
\newcommand{\y}{\bm{y}}
\newcommand{\z}{\bm{z}}
\newcommand{\vtheta}{\bm{\theta}}
\newcommand{\mesh}{\Delta_k^\text{mesh}}
\newcommand{\pollmesh}{\Delta_k^\text{poll}}
\newcommand{\meshz}{\Delta_0^\text{mesh}}
\newcommand{\pollmeshz}{\Delta_0^\text{poll}}
\newcommand{\X}{\mathcal{X}}
\newcommand{\D}{\text{\textbf{D}}}
\newcommand{\I}{{\text{\textbf{I}}}}
\newcommand{\K}{\text{\textbf{K}}}

\newcommand{\nparams}{D}

\newcommand{\var}{s^2}
\newcommand{\sd}{s}
\newcommand{\sigmaest}{\sigma_{\text{est}}}
\newcommand{\ymin}{y_{\text{min}}}
\newcommand{\ybest}{y_{\text{best}}}

\newcommand{\alphahedge}{\alpha_{\text{H}}}
\newcommand{\betahedge}{\beta_{\text{H}}}
\newcommand{\gammahedge}{\gamma_{\text{H}}}

\newcommand{\poll}{\textsc{poll}}
\newcommand{\search}{\textsc{search}}
\newcommand{\supplement}{Supplementary Material}
\newcommand{\ccn}{\textsc{ccn17}}
\newcommand{\bbob}{\textsc{bbob09}}

\newcommand{\xmark}{\ding{55}}%
\newcommand{\maybemark}{$\approx$}%

\newcommand{\norm}[1]{{\lvert\lvert {#1} \rvert\rvert}}

\algnewcommand\algorithmicinput{\textbf{Input:}}
\algnewcommand\INPUT{\item[\algorithmicinput]}

\algloop{Do}

\title{Practical Bayesian Optimization for Model Fitting \\ with Bayesian Adaptive Direct Search}

\author{
Luigi Acerbi\thanks{Current address: 
D\'epartement des neurosciences fondamentales, Universit\'e de Gen\`eve, 
CMU, 1 rue Michel-Servet, 1206 Gen\`eve, Switzerland. E-mail: \texttt{luigi.acerbi@gmail.com}.} 
\\
Center for Neural Science  \\
New York University \\
\texttt{luigi.acerbi@nyu.edu}
 \\
\And
Wei Ji Ma \\
Center for Neural Science \&  Dept. of Psychology \\
New York University \\
\texttt{weijima@nyu.edu}
}

\begin{document}

\maketitle
\vspace{-1em}

\begin{abstract}
\vspace{-0.25em}
Computational models in fields such as computational neuroscience  are often evaluated via stochastic simulation or numerical approximation. Fitting these models implies a difficult optimization problem over complex, possibly noisy parameter landscapes.
Bayesian optimization (BO) has been successfully applied to solving expensive black-box problems in engineering and machine learning. 
Here we explore whether BO can be applied as a general tool for model fitting.
First, we present a novel hybrid BO algorithm, Bayesian adaptive direct search (BADS), that achieves competitive performance with an affordable computational overhead for the running time of typical models.
We then perform an extensive benchmark of BADS vs. many common and state-of-the-art nonconvex, derivative-free optimizers, on a set of model-fitting problems with real data and models from six studies in behavioral, cognitive, and computational neuroscience. With default settings, BADS consistently finds comparable or better solutions than other methods, including `vanilla' BO, showing great promise for advanced BO techniques, and BADS in particular, as a general model-fitting tool.
\end{abstract}

\setcounter{footnote}{0}

\vspace{-0.5em}
\section{Introduction}

\vspace{-0.25em}
Many complex, nonlinear computational models in fields such as behaviorial, cognitive, and computational neuroscience cannot be evaluated analytically, but require moderately expensive numerical approximations or simulations.
% For example, common Bayesian models of cue combination become non-analytical in the presence of stimulus-dependent noise, requiring numerical integration. Models of categorization with high-dimensional stimuli, such as words or objects, require marginalization over features, which can only be approximated via Monte Carlo simulation.
In these cases, finding the maximum-likelihood (ML) solution -- for parameter estimation, or model selection -- requires the costly exploration of a rough or noisy nonconvex landscape, in which gradients are often unavailable to guide the search.

Here we consider the problem of finding the (global) optimum $\x^* = \text{argmin}_{\x \in \X} \mathbb{E}\left[f(\x)\right]$ of a possibly noisy \emph{objective} $f$ over a (bounded) domain $\X \subseteq \mathbb{R}^{\nparams}$, where the function $f$ can be intended as the (negative) log likelihood of a parameter vector $\x$ for a given dataset and model, but is generally a \emph{black box}.
With many derivative-free optimization algorithms available to the researcher \cite{rios2013derivative}, it is unclear which one should be chosen. Crucially, an inadequate optimizer can hinder progress, limit the complexity of the models that can be fit, and even cast doubt on the reliability of one's findings.

\emph{Bayesian optimization} (BO) is a state-of-the-art machine learning framework for optimizing expensive and possibly noisy black-box functions \cite{jones1998efficient,brochu2010tutorial,shahriari2016taking}.
This makes it an ideal candidate for solving difficult model-fitting problems.
Yet there are several obstacles to a widespread usage of BO as a general tool for model fitting. First, traditional BO methods target \emph{very} costly problems, such as hyperparameter tuning \cite{snoek2012practical}, whereas evaluating a typical behavioral model might only have a moderate computational cost (e.g., 0.1-10 s per evaluation). 
This implies major differences in what is considered an acceptable algorithmic overhead, and in the maximum number of allowed function evaluations (e.g., hundreds vs. thousands). Second, it is unclear how BO methods would fare in this regime against commonly used and state-of-the-art, non-Bayesian optimizers. Finally, BO might be perceived by non-practitioners as an advanced tool that requires specific technical knowledge to be implemented or tuned.

We address these issues by developing a novel hybrid BO algorithm, Bayesian Adaptive Direct Search (BADS), that achieves competitive performance at a small computational cost. We tested BADS, together with a wide array of commonly used optimizers, on a novel benchmark set of model-fitting problems with real data and models drawn from studies in cognitive, behaviorial and computational neuroscience. Finally, we make BADS available as a free MATLAB package with the same user interface as existing optimizers and that can be used out-of-the-box with no tuning.\footnote{Code available at \url{https://github.com/lacerbi/bads}.}

BADS is a \emph{hybrid} BO method in that it combines the mesh adaptive direct search (MADS) framework \cite{audet2006mesh} (Section \ref{sec:mads}) with a BO search performed via a local Gaussian process (GP) surrogate (Section \ref{sec:bo}), implemented via a number of heuristics for efficiency  (Section \ref{sec:bads}). BADS proves to be highly competitive on both artificial functions and real-world model-fitting problems (Section \ref{sec:experiments}), showing promise as a general tool for model fitting in computational neuroscience and related fields.

\vspace{-0.5em}
\paragraph{Related work}

There is a large literature about (Bayesian) optimization of expensive, possibly stochastic, computer simulations, mostly used in machine learning \cite{brochu2010tutorial,shahriari2016taking,snoek2012practical} or engineering (known as \emph{kriging-based} optimization) \cite{taddy2009bayesian,picheny2014noisy,gramacy2015mesh}.
Recent work has combined MADS with treed GP models for constrained optimization (TGP-MADS \cite{gramacy2015mesh}).
Crucially, these methods have large overheads and may require problem-specific tuning, making them impractical as a generic tool for model fitting.
Cheaper but less precise surrogate models than GPs have been proposed, such as random forests \cite{hutter2011sequential}, Parzen estimators \cite{bergstra2011algorithms}, and dynamic trees \cite{TaLeDKo2014}. In this paper, we focus on BO based on traditional GP surrogates, leaving the analysis of alternative models for future work (see Conclusions).

\section{Optimization frameworks}

\vspace{-0.25em}
\subsection{Mesh adaptive direct search (MADS)}
\label{sec:mads}

The MADS algorithm is a directional direct search framework for nonlinear optimization \cite{audet2006mesh,audet2008erratum}.
Briefly, MADS seeks to improve the current solution by testing points in the neighborhood of the current point (the \emph{incumbent}), by moving one step in each direction on an iteration-dependent mesh. In addition, the MADS framework can incorporate in the optimization any arbitrary search strategy which proposes additional test points that lie on the mesh.

MADS defines the current mesh at the $k$-th iteration as
$M_k = \bigcup_{\x \in S_k} \left\{ \x + \mesh \D \z : \z \in \mathbb{N}^{\nparams}\right\}$,
where $S_k \subset \mathbb{R}^n$ is the set of all points evaluated since the start of the iteration, $\mesh \in \mathbb{R}_+$ is the \emph{mesh size}, and $\D$ is a fixed matrix in $\mathbb{R}^{\nparams \times n_{\D}}$ whose $n_{\D}$ columns represent viable search directions. We choose $\D = \left[\I_\nparams, -\I_\nparams \right]$, where $\I_\nparams$ is the identity matrix in dimension $\nparams$.

Each iteration of MADS comprises of two stages, a \search{} stage and an optional \poll{} stage. The \search{} stage evaluates a finite number of points proposed by a provided search strategy, with the only restriction that the tested points lie on the current mesh. The search strategy is intended to inject problem-specific information in the optimization. In BADS, we exploit the freedom of \search{} to perform Bayesian optimization in the neighborhood of the incumbent (see Section \ref{sec:bo} and \ref{sec:bmads}).
The \poll{} stage is performed if the \search{} fails in finding a point with an improved objective value. \poll{} constructs a \emph{poll set} of candidate points, $P_k$, defined as
$P_k = \left\{\x_k + \mesh \bm{v} : \bm{v} \in \D_k \right\},$
where $\x_k$ is the incumbent and $\D_k$ is the set of \emph{polling directions} constructed by taking discrete linear combinations of the set of directions $\D$. 
% The set $\D_k$ can be built deterministically or randomly, depending on the MADS implementation. 
The \emph{poll size} parameter $\pollmesh \ge \mesh$ defines the maximum length of poll displacement vectors $\mesh \bm{v}$, for $\bm{v} \in \D_k$ (typically, $\pollmesh \approx \mesh \norm{\bm{v}}$).
Points in the poll set can be evaluated in any order, and the \poll{} is opportunistic in that it can be stopped as soon as a better solution is found.
The \poll{} stage ensures theoretical convergence to a local stationary point according to Clarke calculus for nonsmooth functions \cite{audet2006mesh,clarke1983optimization}. 

If either \search{} or \poll{} are a \emph{success}, finding a mesh point with an improved objective value, the incumbent is updated and the mesh size remains the same or is multiplied by a factor $\tau > 1$. 
If neither  \search{} or \poll{} are successful, the incumbent does not move and the mesh size is divided by $\tau$. The algorithm proceeds until a stopping criterion  is met (e.g., maximum budget of function evaluations).

\vspace{-0.25em}
\subsection{Bayesian optimization}
\label{sec:bo}

The typical form of Bayesian optimization (BO) \cite{jones1998efficient} builds a Gaussian process (GP) approximation of the objective $f$, which is used as a relatively inexpensive surrogate to guide the search towards regions that are promising (low GP mean) and/or unknown (high GP uncertainty), according to a rule, the acquisition function, that formalizes the exploitation-exploration trade-off.

\vspace{-0.5em}
\paragraph{Gaussian processes}

GPs are a flexible class of models for specifying prior distributions over unknown functions $f : \X \subseteq \mathbb{R}^{\nparams} \rightarrow \mathbb{R}$ \cite{rasmussen2006gaussian}. 
GPs are specified by a mean function $m: \X \rightarrow \mathbb{R}$  and a positive definite covariance, or kernel function $k: \X \times \X \rightarrow \mathbb{R}$.
Given any finite collection of $n$ points $\xx = \left\{\xii \in \X \right\}_{i=1}^n$, the value of $f$ at these points is assumed to be jointly Gaussian with mean $(m(\xone),\ldots, m(\xn))^\top$ and covariance matrix $\K$, where $\K_{ij} = k(\xii,\xj)$ for $1 \le i, j \le n$.
We assume i.i.d. Gaussian observation noise such that $f$ evaluated at $\xii$ returns $\yii \sim \mathcal{N}\left(f(\xii), \sigma^2\right)$, and $\y = (\yone,\ldots,\yn )^\top$ is the vector of observed values. For a deterministic $f$, we still assume a small $\sigma > 0$ to improve numerical stability of the GP \cite{gramacy2012cases}.
Conveniently, observation of such (noisy) function values will produce a GP posterior whose latent marginal conditional mean $\mu(\x; \left\{\xx, \y \right\}, \vtheta)$ and variance $\var(\x; \left\{\xx, \y \right\}, \vtheta)$ at a given point are available in closed form (see \supplement{}), where $\vtheta$ is a hyperparameter vector for the mean, covariance, and likelihood.
In the following, we omit the dependency of $\mu$ and $\var$ from the data and GP parameters to reduce clutter.

\vspace{-0.5em}
\paragraph{Covariance functions}

Our main choice of stationary (translationally-invariant) covariance function is the automatic relevance determination (ARD) \emph{rational quadratic} (RQ) kernel,
\begin{equation} \label{eq:rq}
k_\text{RQ}\left(\x,\x^\prime\right) = \sigma_f^2\left[1 + \frac{1}{2\alpha} r^2(\x,\x^\prime) \right]^{-\alpha},
\qquad \text{with} \quad r^2(\x,\x^\prime) = \sum_{d = 1}^{\nparams} \frac{1}{\ell_d^2} \left(x_d - x_d^\prime \right)^2, %  \text{($\bm{\ell}$-scaled distance)}
\end{equation}
where $\sigma_f^2$ is the signal variance, $\ell_1, \ldots, \ell_\nparams$ are the kernel length scales along each coordinate direction, and $\alpha > 0$ is the shape parameter.
More common choices for Bayesian optimization include the \emph{squared exponential} (SE) kernel \cite{gramacy2015mesh} or the twice-differentiable ARD \emph{Mat\'ern 5/2} (M$_{5/2}$) kernel \cite{snoek2012practical}, but we found the RQ kernel to work best in combination with our method (see Section \ref{sec:bbob09}). We also consider \emph{composite periodic kernels} for circular or periodic variables (see \supplement{}).

\vspace{-0.5em}
\paragraph{Acquisition function}

For a given GP approximation of $f$, the \emph{acquisition function}, $a: \X \rightarrow \mathbb{R}$, determines which point in $\X$ should be evaluated next via a proxy optimization $\x_\text{next} = \text{argmin}_{\x} a(\x)$. 
We consider here the \emph{GP lower confidence bound} (LCB) metric \cite{srinivas2010gaussian},
\begin{equation} \label{eq:acqlcb}
a_\text{LCB}\left(\x; \left\{\xx, \y \right\}, \vtheta\right) = \mu\left(\x\right) - \sqrt{\nu \beta_t \var \left(\x\right)}, \qquad \beta_t = 2\ln\left(\nparams t^2 \pi^2/(6 \delta)\right)
\end{equation}
where $\nu > 0$ is a tunable parameter, $t$ is the number of function evaluations so far, $\delta > 0$ is a probabilistic tolerance, and $\beta_t$ is a learning rate chosen to minimize cumulative regret under certain assumptions. For BADS we use the recommended values $\nu = 0.2$ and $\delta = 0.1$ \cite{srinivas2010gaussian}. Another popular choice is the (negative) \emph{expected improvement} (EI) over the current best function value \cite{mockus1978application}, and an historical, less used metric is the (negative) \emph{probability of improvement} (PI) \cite{kushner1964new}.

% We impose an empirical Bayes prior (see Supplementary Material), and approximate the posterior over hyperparameters with a point estimate, or more in general with a set of $\nhyp$ particles computed via Stein variational gradient descent (SVGD) \cite{liu2016stein}. SVGD is a variational inference algorithm that iteratively transports a set of particles to match the target posterior, by applying a form of functional gradient descent that minimizes the KL divergence. In the special case $\nhyp = 1$, SVGD is equivalent to maximum-a-posteriori (MAP) estimation. In practice, we found that a point estimate performs well in our applications, and is compatible with our requirement of fast computations.

\section{Bayesian adaptive direct search (BADS)}
\label{sec:bads}

We describe here the main steps of BADS (Algorithm \ref{alg:bads}).  Briefly, BADS alternates between a series of fast, local BO steps (the \search{} stage of MADS) and a systematic, slower exploration of the mesh grid (\poll{} stage).
The two stages complement each other, in that the \search{} can explore the space very effectively, provided an adequate surrogate model. When the \search{} repeatedly fails, meaning that the GP model is not helping the optimization (e.g., due to a misspecified model, or excess uncertainty), BADS switches to \poll{}. The \poll{} stage performs a fail-safe, model-free optimization, during which BADS gathers information about the local shape of the objective function, so as to build a better surrogate for the next \search{}. This alternation makes BADS able to deal effectively and robustly with a variety of problems. See \supplement{} for a full description.

\subsection{Initial setup}
\label{sec:setup}

\paragraph{Problem specification}

The algorithm is initialized by providing a starting point $\x_0$, vectors of \emph{hard} lower/upper bounds \texttt{LB}, \texttt{UB}, and optional vectors of \emph{plausible} lower/upper bounds \texttt{PLB}, \texttt{PUB}, with the requirement that for each dimension $1\le d\le \nparams$, $\texttt{LB}_d \le \texttt{PLB}_d < \texttt{PUB}_d \le \texttt{UB}_d$.\footnote{A variable $d$ can be \emph{fixed} by setting $(\x_0)_d = \texttt{LB}_d = \texttt{UB}_d = \texttt{PLB}_d = \texttt{PUB}_d$. Fixed variables become constants, and BADS runs on an optimization problem with reduced dimensionality.} Plausible bounds identify a region in parameter space where most solutions are expected to lie. Hard upper/lower bounds can be infinite, but plausible bounds need to be finite. Problem variables whose hard bounds are strictly positive and $\texttt{UB}_d \ge 10 \cdot \texttt{LB}_d$ are automatically converted to log space. All variables are then linearly rescaled to the standardized box $[-1,1]^\nparams$ such that the box bounds correspond to $\left[\texttt{PLB}, \texttt{PUB}\right]$ in the original space. 
BADS supports bound or no constraints, and optionally other constraints via a provided \emph{barrier} function $c$ (see \supplement{}). 
The user can also specify circular or periodic dimensions (such as angles); and whether the objective $f$ is deterministic or noisy (stochastic), and in the latter case provide a coarse estimate of the noise (see Section \ref{sec:noisy}).

\vspace{-0.5em}
\paragraph{Initial design}

The initial design consists of the provided starting point $\x_0$ and $n_{\text{init}} = \nparams$ additional points chosen via a space-filling quasi-random Sobol sequence \cite{bratley1988algorithm} in the standardized box, and forced to lie on the mesh grid. 
If the user does not specify whether $f$ is deterministic or stochastic, the algorithm assesses it by performing two consecutive evaluations at $\x_0$.

\begin{algorithm}[t]
\caption{Bayesian Adaptive Direct Search}\label{alg:bads}
\begin{algorithmic}[1]
\INPUT objective function $f$, starting point $\bm{\x_0}$, hard bounds \texttt{LB}, \texttt{UB}, (\emph{optional}: plausible bounds \texttt{PLB}, \texttt{PUB}, barrier function $c$, additional \texttt{options})
\State \textbf{Initialization:} 
$\meshz \leftarrow 2^{-10}$, $\pollmeshz \leftarrow 1$, $k \leftarrow 0$, evaluate $f$ on initial design \Comment{Section \ref{sec:setup}}
\Repeat
\State (update GP approximation at any step; refit hyperparameters if necessary) \Comment{Section \ref{sec:gp}}
\For{$1 \ldots n_\text{search}$}  \Comment{\search{} stage, Section \ref{sec:bmads}}% $n_\text{search} = \max\{\nparams, \lfloor 3 + \nparams/2\rfloor \}$}
\State $\xsearch \leftarrow $ \textsc{SearchOracle} \Comment{local Bayesian optimization step}
\State Evaluate $f$ on $\xsearch$, \textbf{if} improvement is \emph{sufficient} \textbf{then break}
\EndFor
\If{\textsc{search} is \textsc{not} \emph{successful}}  \Comment{optional \poll{} stage, Section \ref{sec:bmads}}
\State compute poll set $P_k$
\State evaluate opportunistically $f$ on $P_k$ sorted by acquisition function
\EndIf
\If{iteration $k$ is \emph{successful}}
\State update incumbent $\x_{k + 1}$
\State \textbf{if} \poll{} was \emph{successful} \textbf{then} $\mesh \leftarrow 2 \mesh$, $\pollmesh \leftarrow 2 \pollmesh$ 
\Else
\State $\mesh \leftarrow \frac{1}{2} \mesh$, $\pollmesh \leftarrow \frac{1}{2} \pollmesh$
\EndIf
\State $k \leftarrow k + 1$
\Until{\texttt{fevals} $>$ \texttt{MaxFunEvals} \textbf{ or } $\pollmesh < 10^{-6} $ \textbf{ or } stalling} \Comment{stopping criteria}
\State \Return $\x_{\text{end}} = \arg \min_k f(\x_k)$ (or $\x_{\text{end}} = \arg \min_k q_{\beta}(\x_k)$ for noisy objectives,  Section \ref{sec:noisy})
\end{algorithmic}
\end{algorithm}

\subsection{GP model in BADS}
\label{sec:gp}

The default GP model is specified by a constant mean function $m \in \mathbb{R}$, a smooth ARD RQ kernel (Eq. \ref{eq:rq}), and we use $a_\text{LCB}$ (Eq. \ref{eq:acqlcb}) as a default acquisition function.

\vspace{-0.5em}
\paragraph{Hyperparameters} The default GP has hyperparameters $\vtheta = (\ell_1, \ldots, \ell_\nparams, \sigma_f^2, \alpha, \sigma^2, m)$.
We impose an empirical Bayes prior on the GP hyperparameters based on the current training set (see Supplementary Material), and select $\vtheta$ via maximum a posteriori (MAP) estimation. We fit $\vtheta$ via a gradient-based nonlinear optimizer, 
starting from either the previous value of $\vtheta$ or a weighted draw from the prior, as a means to escape local optima. We refit the hyperparameters every $2 \nparams$ to $5 \nparams$ function evaluations; more often earlier in the optimization, and whenever the current GP is particularly inaccurate at predicting new points, according to a normality test on the residuals, $\zii = \left(\yii - \mu(\xii)\right)/\sqrt{\var(\xii) + \sigma^2}$ (assumed independent, in first approximation).

\vspace{-0.5em}
\paragraph{Training set} The GP training set $\xx$ consists of a subset of the points evaluated so far (the \emph{cache}), selected to build a local approximation of the objective in the neighborhood of the incumbent $\x_k$, constructed as follows. Each time $\xx$ is rebuilt, points in the cache are sorted by their $\bm{\ell}$-scaled distance $r^2$ (Eq. \ref{eq:rq}) from $\x_k$.
First, the closest $n_\text{min} = 50$ points are automatically added to $\xx$. Second, up to $10 \nparams$ additional points with $r \le 3 \rho(\alpha)$ are included in the set, where $\rho(\alpha) \gtrsim 1$ is a radius function that depends on the decay of the kernel. For the RQ kernel, $\rho_\text{RQ}(\alpha) = \sqrt{\alpha} \sqrt{e^{1/\alpha} - 1}$
(see Supplementary Material). 
% Finally, for any $1\le d\le \nparams$ we make sure to add to the training set \emph{safeguard points} $\x^\prime$, $\x^{\prime\prime}$ (if they exist in the cache) such that ${x^\prime}^{(d)} < x_k^{(d)}$ and ${x^{\prime\prime}}^{(d)} > x_k^{(d)}$. 
Newly evaluated points are added incrementally to the set, using fast rank-one updates of the GP posterior. The training set is rebuilt any time the incumbent is moved.

\subsection{Implementation of the MADS framework}
\label{sec:bmads}

We initialize $\pollmeshz = 1$ and $\meshz =  2^{-10}$ (in standardized space), such that the initial poll steps can span the plausible region, whereas the mesh grid is relatively fine. We use $\tau = 2$, and increase the mesh size only after a successful \poll{}. We skip the \poll{} after a successful \search{}.

\vspace{-0.5em}
\paragraph{Search stage} 
We apply an aggressive, repeated \search{} strategy that consists of up to $n_{\text{search}} = \max\{\nparams,\lfloor3+\nparams/2\rfloor\}$ unsuccessful \search{} steps. In each step, we use a \emph{search oracle}, based on a local BO with the current GP, 
to produce a search point $\xsearch$ (see below). We evaluate $f(\xsearch)$ and add it to the training set. If the improvement in objective value is none or \emph{insufficient}, that is less than $(\pollmesh)^{3/2}$, we continue searching, or switch to \poll{} after $n_{\text{search}}$ steps. Otherwise, we call it a \emph{success} and start a new \search{} from scratch, centered on the updated incumbent.

\vspace{-0.5em}
\paragraph{Search oracle} 
We choose $\xsearch$ via a fast, approximate optimization inspired by CMA-ES \cite{hansen2003reducing}. We sample batches of points in the neighborhood of the incumbent $\x_k$, drawn $\sim \mathcal{N}(\x_\text{s}, \lambda^2 (\pollmesh)^2 \bm{\Sigma})$, where $\x_\text{s}$ is the current search focus, $\bm{\Sigma}$ a \emph{search covariance matrix}, and $\lambda > 0$ a scaling factor, and we pick the point that optimizes the acquisition function (see \supplement{}).
We remove from the \search{} set candidate points that violate non-bound constraints ($c(\x) > 0$), and we project candidate points that fall outside hard bounds to the closest mesh point inside the bounds.
Across \search{} steps, we use both a diagonal matrix $\bm{\Sigma}_{\bm{\ell}}$ with diagonal  $\left(\ell_1^2 / |\bm{\ell}|^2, \ldots, \ell_{\nparams}^2 / |\bm{\ell}|^2\right)$, and a matrix $\bm{\Sigma}_{\text{WCM}}$ proportional to the weighted covariance matrix of points in $\xx$ (each point weighted according to a function of its ranking in terms of objective values $y_i$). We choose between $\bm{\Sigma}_{\bm{\ell}}$ and $\bm{\Sigma}_{\text{WCM}}$ probabilistically via a \emph{hedge} strategy, based on their track record of cumulative improvement \cite{hoffman2011portfolio}.

\vspace{-0.5em}
\paragraph{Poll stage} 
We incorporate the GP approximation in the \poll{} in two ways: when constructing the set of polling directions $\D_k$, and when choosing the polling order. 
We generate $\D_k$ according to the random LTMADS algorithm \cite{audet2006mesh}, but then rescale each vector coordinate $1 \le d \le \nparams$ proportionally to the GP length scale $\ell_d$ (see \supplement{}). We discard poll vectors that do not satisfy the given bound or nonbound constraints.
Second, since the \poll{} is opportunistic, we evaluate points in the poll set according to the ranking given by the acquisition function \cite{gramacy2015mesh}.

\vspace{-0.5em}
\paragraph{Stopping criteria} We stop the optimization when the poll size $\pollmesh$ goes below a threshold (default $10^{-6}$); 
when reaching a maximum number of objective evaluations (default $500 \nparams$); or if there is no significant improvement of the objective for more than $4 + \lfloor \nparams/2 \rfloor$ iterations. The algorithm returns the optimum $\x_{\text{end}}$ (transformed back to original coordinates) with the lowest objective value $y_{\text{end}}$.

\subsection{Noisy objective}
\label{sec:noisy}

In case of a noisy objective, we assume for the noise a hyperprior  $\ln \sigma \sim \mathcal{N}(\ln \sigmaest, 1)$, with $\sigmaest$ a base noise magnitude (default $\sigmaest = 1$, but the user can provide an estimate). 
To account for additional uncertainty, we also make the following changes: double the minimum number of points added to the training set, $n_\text{min} = 100$, and increase the maximum number to 200; increase the initial design to $n_{\text{init}} = 20$; and double the number of allowed  stalled iterations before stopping.

\vspace{-0.5em}
\paragraph{Uncertainty handling}
Due to noise, we cannot simply use the output values $y_i$ as ground truth in the \search{} and \poll{} stages. Instead, we replace $y_i$ with the GP latent quantile function \cite{picheny2013quantile}
\begin{equation} \label{eq:quantiles}
q_\beta\left(\x; \left\{\xx, \y \right\}, \vtheta\right) \equiv q_\beta(\x) = \mu\left(\x\right) + \Phi^{-1}(\beta) \sd\left(\x\right), \qquad \beta \in [0.5,1),
\end{equation}
where $\Phi^{-1}(\cdot)$ is the quantile function of the standard normal (\emph{plugin} approach \cite{picheny2013benchmark}).
Moreover, we modify the MADS procedure by keeping an \emph{incumbent set} 
$\{\x_i\}_{i=1}^k$, where $\x_i$ is the incumbent at the end of the $i$-th iteration. At the end of each \poll{} we re-evaluate $q_\beta$ for all elements of the incumbent set, in light of the new points added to the cache. We select as current (active) incumbent the point with lowest $q_\beta(\x_i)$. During optimization we set $\beta = 0.5$ (mean prediction only), which promotes exploration. We use a conservative $\beta_\text{end} = 0.999$ for the last iteration, to select the optimum $\x_{\text{end}}$ returned by the algorithm in a robust manner. 
Instead of $y_\text{end}$, we return either $\mu(\x_{\text{end}})$ or an unbiased estimate of $\mathbb{E}[f(\x_\text{end})]$ obtained by averaging multiple evaluations (see \supplement{}).

\section{Experiments}
\label{sec:experiments}

We tested BADS and many optimizers with implementation available in MATLAB (R2015b, R2017a) on a large set of artificial and real optimization problems (see \supplement{} for details).

\vspace{-0.25em}
\subsection{Design of the benchmark}
\label{sec:design}

\paragraph{Algorithms}

Besides BADS, we tested 16 optimization algorithms, including popular choices such as Nelder-Mead (\texttt{fminsearch} \cite{lagarias1998convergence}), several constrained nonlinear optimizers in the \texttt{fmincon} function (default \emph{interior-point} \cite{waltz2006interior}, 
\emph{sequential quadratic programming} \texttt{sqp} \cite{wright1999numerical}, and \emph{active-set} \texttt{actset} \cite{gill1981practical}), genetic algorithms (\texttt{ga} \cite{goldberg1989genetic}), random search (\texttt{randsearch}) as a baseline \cite{bergstra2012random}; and also less-known state-of-the-art methods for nonconvex derivative-free optimization \cite{rios2013derivative}, such as Multilevel Coordinate Search (MCS \cite{huyer1999global}) and CMA-ES \cite{hansen2003reducing,jastrebski2006improving} (\texttt{cmaes}, in different flavors). For noisy objectives, we included algorithms that explicitly handle uncertainty, such as \texttt{snobfit} \cite{csendes2008global} and \emph{noisy} CMA-ES \cite{hansen2009method}.
Finally, to verify the advantage of BADS' hybrid approach to BO, we also tested a standard, `vanilla' version of BO \cite{snoek2012practical} (\texttt{bayesopt}, R2017a) on the set of real model-fitting problems (see below).
For all algorithms, including BADS, we used default settings (no fine-tuning).

\vspace{-0.5em}
\paragraph{Problem sets}

First, we considered a standard benchmark set of artificial, noiseless functions (\bbob{} \cite{hansen2009real}, 24 functions) in dimensions $\nparams \in \{3,6,10,15 \}$, for a total of $96$ test functions. We also created `noisy' versions of the same set.
Second, we collected model-fitting problems from six published or ongoing studies in cognitive and computational neuroscience (\ccn{}). The objectives of the \ccn{} set are negative log likelihood functions of an input parameter vector, for specified datasets and models, and can be deterministic or stochastic.
For each study in the \ccn{} set we asked its authors for six different real datasets (i.e., subjects or neurons), divided between one or two main models of interest; collecting a total of 36 test functions with $D \in \{6,9,10,12,13 \}$.

\vspace{-0.5em}
\paragraph{Procedure}

We ran 50 independent runs of each algorithm on each test function, with randomized starting points and a budget of $500 \times D$ function evaluations ($200 \times D$ for noisy problems). If an algorithm terminated before depleting the budget, it was restarted from a new random point.
We consider a run \emph{successful} if the current best (or returned, for noisy problems) function value is within a given \emph{error tolerance} $\varepsilon > 0$ from the true optimum $f_\text{min}$ (or our best estimate thereof).\footnote{Note that the error tolerance $\varepsilon$ is \emph{not} a fractional error, as sometimes reported in optimization, because for model comparison we typically care about (absolute) differences in log likelihoods.}
For noiseless problems, we compute the fraction of successful runs as a function of number of objective evaluations, averaged over datasets/functions and over $\varepsilon \in [0.01, 10]$ (log spaced). This is a realistic range for $\varepsilon$, as differences in log likelihood below 0.01 are irrelevant for model selection; an acceptable tolerance is $\varepsilon \sim 0.5$ (a difference in \emph{deviance}, the metric used for AIC or BIC, less than 1); larger $\varepsilon$ associate with coarse solutions, but errors larger than 10 would induce excessive biases in model selection.  For noisy problems, what matters most is the solution $\x_{\text{end}}$ that the algorithm \emph{actually} returns, which, depending on the algorithm, may not necessarily be the point with the lowest \emph{observed} function value.
Since, unlike the noiseless case, we generally do not know the solutions that would be returned by any algorithm at every time step, but only at the last step, we plot instead the fraction of successful runs at $200 \times D$ function evaluations as a function of $\varepsilon$, for $\varepsilon \in [0.1, 10]$ (noise makes higher precisions moot), and averaged over datasets/functions.
In all plots we omit error bars for clarity (standard errors would be about the size of the line markers or less).

\subsection{Results on artificial functions (\bbob{})}
\label{sec:bbob09}

\begin{figure}[htb]
  \includegraphics[width=\linewidth]{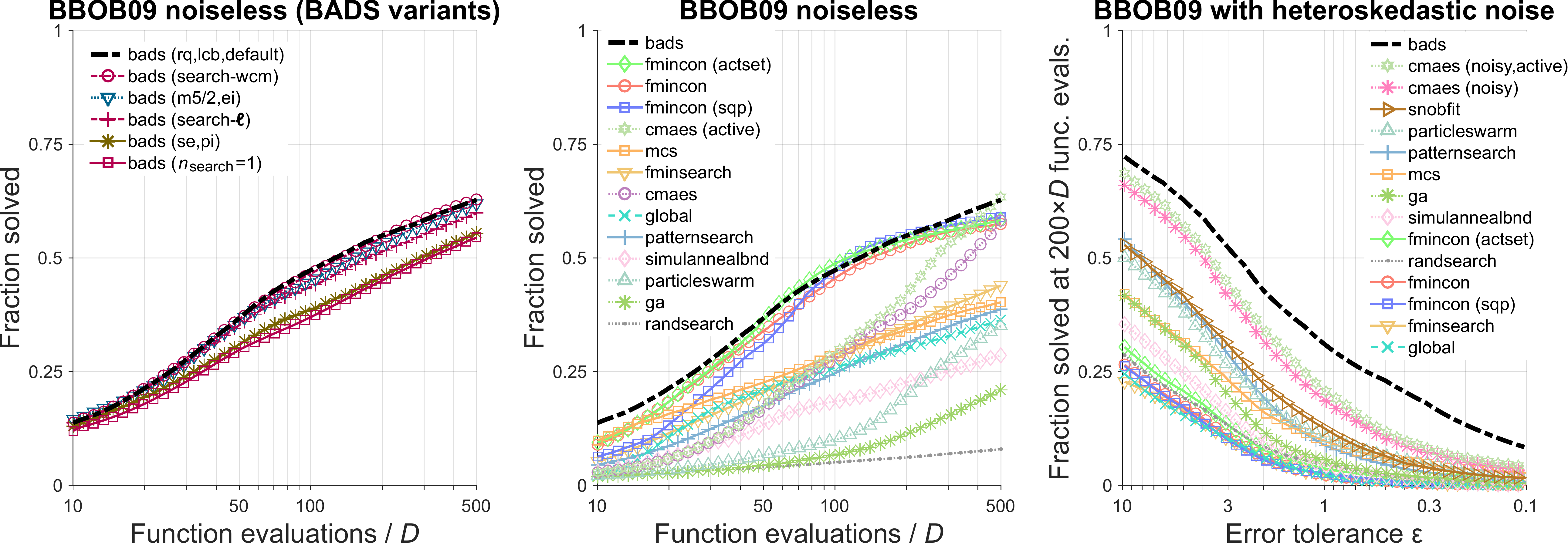}  
\vspace{-1.75em}
  \caption{{\bf Artificial test functions (\bbob{}).} \emph{Left \& middle}: Noiseless functions. Fraction of successful runs ($\varepsilon \in [0.01,10]$) vs. \# function evaluations per \# dimensions, for $\nparams \in \{3,6,10,15 \}$ (96 test functions); for different BADS configurations (\emph{left}) and all algorithms (\emph{middle}). \emph{Right}: Heteroskedastic noise. Fraction of successful runs at $200\times \nparams$ objective evaluations vs. tolerance $\varepsilon$.}
  \label{fig:bbob09}
\end{figure}

The \bbob{} noiseless set \cite{hansen2009real} comprises of 24 functions divided in 5 groups with different properties: separable; low or moderate conditioning; unimodal with high conditioning; multi-modal with adequate / with weak global structure.
% separable ($f_1$-$f_5$); low or moderate conditioning ($f_6$-$f_9$); unimodal with high conditioning ($f_{10}$-$f_{14}$); multi-modal with adequate global structure ($f_{15}$-$f_{19}$); multi-modal with weak global structure ($f_{20}$-$f_{24}$). 
First, we use this benchmark to show the performance of different configurations for BADS. 
Note that we selected the default configuration (RQ kernel, $a_{\text{LCB}}$) and other algorithmic details by testing on a different benchmark set (see \supplement{}).
Fig \ref{fig:bbob09} (left) shows aggregate results across all noiseless functions with $\nparams \in \{3, 6, 10, 15 \}$, for alternative choices of kernels and acquisition functions (only a subset is shown, such as the popular M$_{5/2}$, EI combination), or by altering other features (such as setting $n_\text{search}=1$, or fixing the search covariance matrix to $\bm{\Sigma}_{\bm{\ell}}$ \emph{or} $\bm{\Sigma}_{\text{WCM}}$).
Almost all changes from the default configuration worsen performance.

\vspace{-0.5em}
\paragraph{Noiseless functions} We then compared BADS to other algorithms (Fig \ref{fig:bbob09} middle). Depending on the number of function evaluations, the best optimizers are~BADS, methods of the \texttt{fmincon} family, and, for large budget of function evaluations, CMA-ES with \emph{active} update of the covariance matrix.

\vspace{-0.5em}
\paragraph{Noisy functions} We produce noisy versions of the \bbob{} set by adding i.i.d. Gaussian observation noise at each function evaluation, $y^{(i)} = f(\x^{(i)}) + \sigma(\x^{(i)}) \eta^{(i)}$, with $\eta^{(i)} \sim \mathcal{N}(0,1)$. We consider a variant with moderate \emph{homoskedastic} (constant) noise ($\sigma = 1$), and a variant with \emph{heteroskedastic} noise with $\sigma(\x) = 1 + 0.1 \times (f(\x)-f_\text{min})$, which follows the observation that variability generally increases for solutions away from the optimum. For many functions in the \bbob{} set, this heteroskedastic noise can become substantial ($\sigma \gg 10$) away from the optimum. Fig \ref{fig:bbob09} (right) shows aggregate results for the heteroskedastic set (homoskedastic results are similar). BADS outperforms all other optimizers, with CMA-ES (\emph{active}, with or without the \emph{noisy} option) coming second.

Notably, BADS performs well even on problems with non-stationary (location-dependent) features, such as heteroskedastic noise, thanks to its local GP approximation.

\subsection{Results on real model-fitting problems (\ccn{})}
\label{sec:ccn}

\begin{figure}[tbh]
  \includegraphics[width=\linewidth]{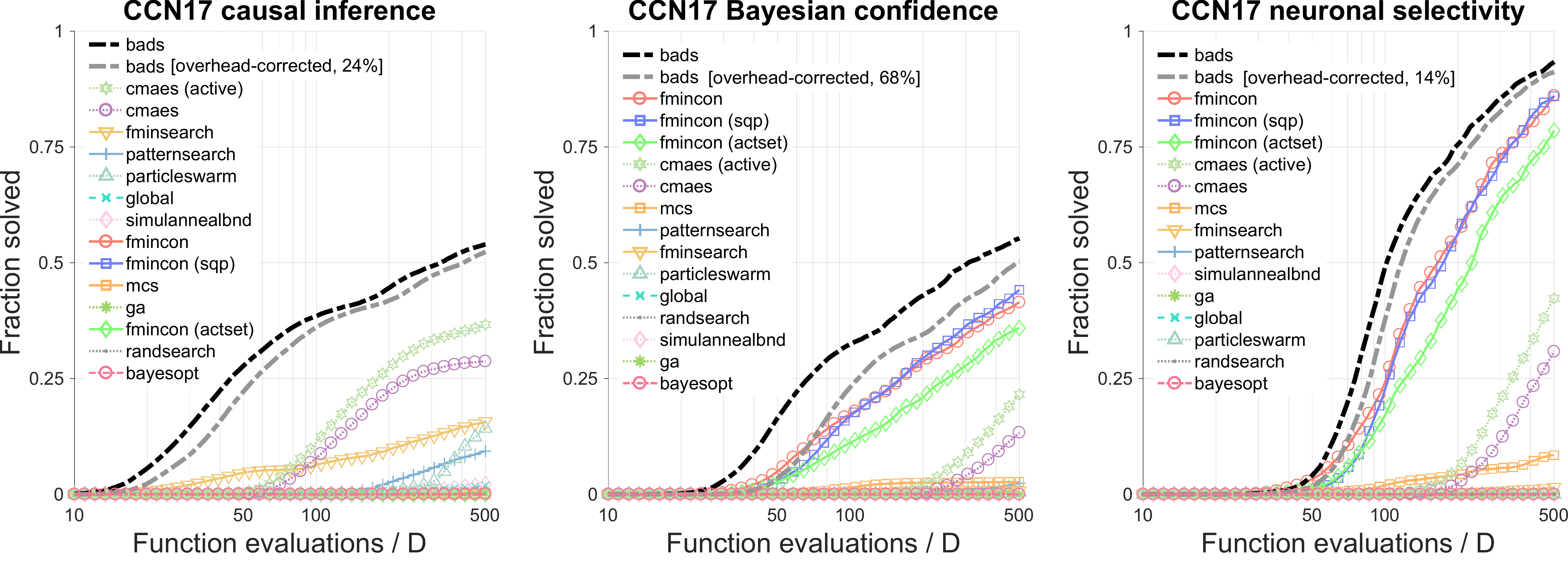}  
  \vspace{-1.75em}
  \caption{{\bf Real model-fitting problems (\ccn{}, deterministic).} Fraction of successful runs ($\varepsilon \in [0.01,10]$) vs. \# function evaluations per \# dimensions. \emph{Left}: Causal inference in visuo-vestibular perception \cite{acerbi2017bayesian} (6 subjects, $\nparams = 10$). \emph{Middle}: Bayesian confidence in perceptual categorization \cite{adler2017human} (6 subjects, $\nparams = 13$). \emph{Right}: Neural model of orientation selectivity \cite{goris2015origin} (6 neurons, $\nparams = 12$). }
  \label{fig:ccn17a}
\end{figure}

\begin{figure}[tb]
  \includegraphics[width=\linewidth]{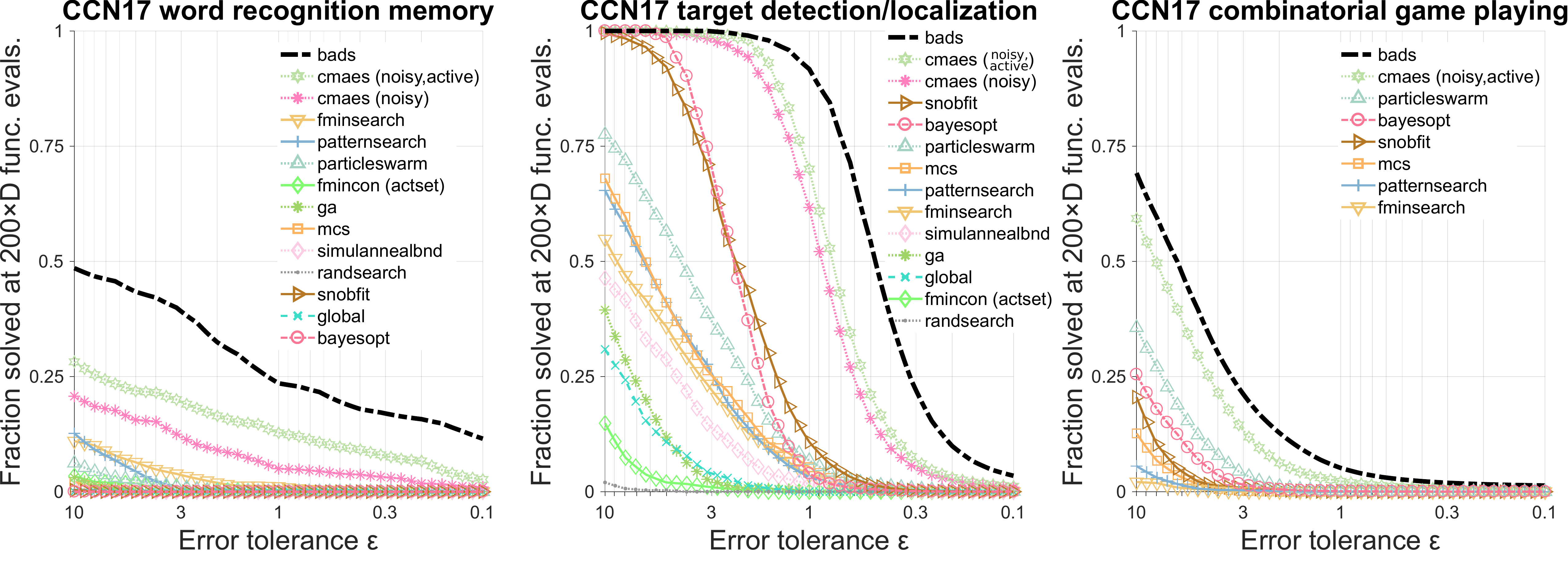}  
\vspace{-1.75em}  
\caption{{\bf Real model-fitting problems (\ccn{}, noisy).} Fraction of successful runs at $200\times \nparams$ objective evaluations vs. tolerance $\varepsilon$. \emph{Left}: Confidence in word recognition memory \cite{van2017fechner} (6 subjects, $\nparams = 6,9$). \emph{Middle}: Target detection and localization \cite{mazyar2012does} (6 subjects, $\nparams = 6$). \emph{Right}: Combinatorial board game playing \cite{van2016people} (6 subjects, $\nparams = 10$). \vspace{-1.1em}}
  \label{fig:ccn17b}
\end{figure}

The objectives of the \ccn{} set are deterministic (e.g., computed via numerical approximation) for three studies (Fig \ref{fig:ccn17a}), and noisy (e.g., evaluated via simulation) for the other three (Fig \ref{fig:ccn17b}). 

The algorithmic cost of BADS is $\sim 0.03$ s to $0.15$ s per function evaluation, depending on $\nparams$, mostly due to the refitting of the GP hyperparameters.
This produces a non-negligible \emph{overhead}, defined as $100\% \, \times$ (\emph{total optimization time} / \emph{total function time} $- 1$).
For a fair comparison with other methods with little or no overhead, for deterministic problems we also plot the \emph{effective} performance of BADS by accounting for the extra cost per function evaluation. In practice, this correction shifts rightward the performance curve of BADS in log-iteration space, since each function evaluation with BADS has an increased fractional time cost.
For stochastic problems, we cannot compute effective performance as easily, but there we found small overheads ($< 5\%$), due to more costly evaluations (more than 1 s).

For a direct comparison with standard BO, we also tested on the \ccn{} set a `vanilla' BO algorithm, as implemented in MATLAB R2017a (\texttt{bayesopt}). This implementation closely follows  \cite{snoek2012practical}, with optimization instead of marginalization over GP hyperparameters.
Due to the fast-growing cost of BO as a function of training set size, we allowed up to 300 training points for the GP, restarting the BO algorithm from scratch with a different initial design every 300 BO iterations (until the total budget of function evaluations was exhausted). The choice of 300 iterations already produced a large average algorithmic overhead of $\sim 8$ s per function evaluation. In showing the results of \texttt{bayesopt}, we display raw performance without penalizing for the overhead.

\paragraph{Causal inference in visuo-vestibular perception}

Causal inference (CI) in perception is the process whereby the brain decides whether to integrate or segregate multisensory cues that could arise from the same or from different sources \cite{kording2007causal}. This study investigates CI in visuo-vestibular heading perception across tasks and under different levels of visual reliability, via a factorial model comparison \cite{acerbi2017bayesian}. For our benchmark we fit three subjects with a Bayesian CI model ($\nparams = 10$), and another three with a fixed-criterion CI model ($\nparams = 10$) that disregards visual reliability. Both models include heading-dependent likelihoods and marginalization of the decision variable over the latent space of noisy sensory measurements $(x_\text{vis}, x_\text{vest})$,  solved via nested numerical integration in 1-D and 2-D.

\vspace{-0.5em}
\paragraph{Bayesian confidence in perceptual categorization}

This study investigates the \emph{Bayesian confidence hypothesis} that subjective judgments of confidence are directly related to the posterior probability the observer assigns to a learnt perceptual category \cite{adler2017human} (e.g., whether the orientation of a drifting Gabor patch belongs to a `narrow' or to a `wide' category). For our benchmark we fit six subjects to the `Ultrastrong' Bayesian confidence model ($\nparams = 13$), which uses the same mapping between posterior probability and confidence across two tasks with different distributions of stimuli. This model includes a latent noisy decision variable, marginalized over via 1-D numerical integration.

\vspace{-0.5em}
\paragraph{Neural model of orientation selectivity}

The authors of this study explore the origins of diversity of neuronal orientation selectivity in visual cortex via novel stimuli (orientation mixtures) and modeling \cite{goris2015origin}. We fit the responses of five V1 and one V2 cells with the authors' neuronal model ($\nparams = 12$) that combines effects of filtering, suppression, and response nonlinearity \cite{goris2015origin}. The model has one circular parameter, the preferred direction of motion of the neuron. The model is analytical but still computationally expensive due to large datasets and a cascade of several nonlinear operations.

\vspace{-0.5em}
\paragraph{Word recognition memory}

This study models a word recognition task in which subjects rated their confidence that a presented word was in a previously studied list \cite{van2017fechner} (data from \cite{mickes2007direct}). 
We consider six subjects divided between two normative models, the `Retrieving Effectively from Memory' model \cite{shiffrin1997model} ($\nparams = 9$) and a similar, novel model\footnote{Unpublished; upcoming work from Aspen H. Yoo and Wei Ji Ma.} ($\nparams = 6$). Both models use Monte Carlo methods to draw random samples from a large space of latent noisy memories, yielding a stochastic log likelihood.

\vspace{-0.5em}
\paragraph{Target detection and localization} 

This study looks at differences in observers' decision making strategies in target detection (`was the target present?') and localization (`which one was the target?') with displays of $2,3,4,$ or $6$ oriented Gabor patches.\footnote{Unpublished; upcoming work from Andra Mihali and Wei Ji Ma.} Here we fit six subjects with a previously derived ideal observer model \cite{ma2011behavior,mazyar2012does} ($\nparams = 6$) with variable-precision noise \cite{van2012variability}, assuming shared parameters between detection and localization. The log likelihood is evaluated via simulation due to marginalization over latent noisy measurements of stimuli orientations with variable precision.

\vspace{-0.5em}
\paragraph{Combinatorial board game playing} This study analyzes people's strategies in a four-in-a-row game played on a 4-by-9 board against human opponents (\cite{van2016people}, Experiment 1). We fit the data of six players with the \emph{main} model ($\nparams = 10$), which is based on a Best-First exploration of a decision tree guided by a feature-based value heuristic.
The model also includes feature dropping, value noise, and lapses, to better capture human variability. Model evaluation is computationally expensive due to the construction and evaluation of trees of future board states, and achieved via \emph{inverse binomial sampling}, an unbiased stochastic estimator of the log likelihood \cite{van2016people}. Due to prohibitive computational costs, here we only test major algorithms (MCS is the method used in the paper \cite{van2016people}); see Fig \ref{fig:ccn17b} right.

In all problems, BADS consistently performs on par with or outperforms all other tested optimizers, even when accounting for its extra algorithmic cost. The second best algorithm is either some flavor of CMA-ES or, for some deterministic problems, a member of the \texttt{fmincon} family. Crucially, their ranking across problems is inconsistent, with both CMA-ES and \texttt{fmincon} performing occasionally quite poorly (e.g., \texttt{fmincon} does poorly in the \emph{causal inference} set because of small fluctuations in the log likelihood landscape caused by coarse numerical integration).
Interestingly, vanilla BO (\texttt{bayesopt}) performs poorly on all problems, often at the level of random search, and always 
substantially worse than BADS, even without accounting for the much larger overhead of \texttt{bayesopt}. The solutions found by \texttt{bayesopt} are often hundreds (even thousands) points of log likelihood from the optimum. This failure is possibly due to the difficulty of building a \emph{global} GP surrogate for BO, coupled with strong non-stationarity of the log likelihood functions; and might be ameliorated by more complex forms of BO (e.g., input warping to produce nonstationary kernels \cite{snoek2014input}, hyperparameter marginalization \cite{snoek2012practical}). However, these advanced approaches would substantially increase the already large overhead. Importantly, we expect this poor perfomance to extend to any package which implements vanilla BO (such as \emph{BayesOpt} \cite{martinez2014bayesopt}), regardless of the efficiency of implementation.

\vspace{-0.25em}
\section{Conclusions}

We have developed a novel BO method and an associated toolbox, BADS, with the goal of fitting moderately expensive computational models out-of-the-box. We have shown on real model-fitting problems that BADS outperforms widely used and state-of-the-art methods for nonconvex, derivative-free optimization, including `vanilla' BO.
We attribute the robust performance of BADS to the alternation between the aggressive \search{} strategy, based on local BO, and the failsafe \poll{} stage, which protects against failures of the GP surrogate -- whereas vanilla BO does not have such failsafe mechanisms, and can be strongly affected by model misspecification.
Our results demonstrate that a hybrid Bayesian approach to optimization can be beneficial beyond the domain of very costly black-box functions, in line with recent advancements in probabilistic numerics \cite{hennig2015probabilistic}. 

Like other surrogate-based methods, the performance of BADS is linked to its ability to obtain a fast approximation of the objective, which generally deteriorates in high dimensions, or for functions with pathological structure (often improvable via reparameterization).
From our tests, we recommend BADS, paired with some multi-start optimization strategy, for models with up to $\sim15$ variables,  a noisy or jagged log likelihood landscape, and when algorithmic overhead is $\lesssim75\%$ (e.g., model evaluation $\gtrsim 0.1$ s).
Future work with BADS will focus on testing alternative statistical surrogates instead of GPs \cite{TaLeDKo2014}; combining it with a smart multi-start method for global optimization; providing support for tunable precision of noisy observations \cite{picheny2013quantile}; improving the numerical implementation; and recasting some of its heuristics in terms of approximate inference.

\clearpage

\section*{Acknowledgments}

We thank Will Adler, Robbe Goris, Andra Mihali, Bas van Opheusden,
and Aspen Yoo for sharing data and model evaluation code that we used in the \ccn{} benchmark set; 
Maija Honig, Andra Mihali, Bas van Opheusden, and Aspen Yoo for providing user feedback on earlier versions of the \texttt{bads} package for MATLAB; Will Adler, Andra Mihali, Bas van Opheusden, and Aspen Yoo for helpful feedback on a previous version of this manuscript; John Wixted and colleagues for allowing us to reuse their data for the \ccn{} `word recognition memory' problem set; and three anonymous reviewers for useful feedback. This work has utilized the NYU IT High Performance Computing resources and services.

{\def\section*#1{}
{

}
}

\clearpage

\part*{Supplementary Material}

\appendix

\setcounter{footnote}{0}
\setcounter{figure}{0}
\setcounter{table}{0}
\setcounter{equation}{0}
\renewcommand{\theequation}{S\arabic{equation}}
\renewcommand{\thetable}{S\arabic{table}}
\renewcommand{\thefigure}{S\arabic{figure}}

In this Supplement, we expand on the definitions and implementations of Gaussian Processes (GPs) and Bayesian optimization in BADS (Section \ref{sec:gpbads}); we give a full description of the BADS algorithm, including details omitted in the main text (Section \ref{sec:extrabads}); we report further details of the benchmark procedure, such as the full list of tested algorithms and additional results (Section \ref{sec:benchmark}); and, finally, we briefly discuss the numerical implementation (Section \ref{sec:implementation}).

\section{Gaussian processes for Bayesian optimization in BADS}
\label{sec:gpbads}

In this section, we describe definitions and additional specifications of the Gaussian process (GP) model used for Bayesian optimization (BO) in BADS. Specifically, this part expands on Sections \ref{sec:bo} and \ref{sec:gp} in the main text.

\paragraph{GP posterior moments}

We consider a GP based on a training set $\xx$ with $n$ points, a vector of observed function values $\y$, and GP mean function $m(\x)$ and GP covariance or kernel function $k(\x,\x^\prime)$, with i.i.d. Gaussian observation noise $\sigma^2 > 0$.
The GP posterior latent marginal conditional mean $\mu$ and variance $\var$ are available in closed form at a chosen point as
\begin{equation} \label{eq:gpmoments}
\begin{split}
\mu\left(\x\right) \equiv \mu\left(\x; \left\{\xx, \y \right\}, \vtheta\right) = & \, \k(\x)^\top \left(\K + \sigma^2 \I_n \right)^{-1} \left(\y - m(\x)\right) \\
\var\left(\x\right) \equiv \var\left(\x; \left\{\xx, \y \right\}, \vtheta\right) = & \, k(\x,\x) - \k(\x)^\top \left(\K + \sigma^2 \I_n \right)^{-1} \k(\x) \\
\end{split}
\end{equation}
where $\K_{ij} = k(\xii,\xj)$, for $1 \le i, j \le n$, is the kernel matrix, $\k(\x) \equiv (k(\x, \xone), \ldots, k(\x, \xn))^\top$ is the $n$-dimensional column vector of cross-covariances, and $\vtheta$ is the vector of GP hyperparameters.

\subsection{Covariance functions}
\label{sec:covariance}

Besides the automatic relevance determination (ARD) \emph{rational quadratic} (RQ) kernel described in the main text (and BADS default), we also considered the common \emph{squared exponential} (SE) kernel
\begin{equation} \label{eq:smoothkernel}
k_\text{SE}\left(\x,\x^\prime\right) = \sigma_f^2 \exp \left\{-\frac{1}{2} r^2(\x,\x^\prime)\right\}, 
\qquad \text{with} \; r^2(\x,\x^\prime) = \sum_{d = 1}^{\nparams} \frac{1}{\ell_d^2} \left(x_d - x_d^\prime \right)^2, %  \qquad \qquad \text{($\bm{\ell}$-scaled distance)}
%\qquad k_\text{RQ}\left(\x,\x^\prime\right) = \sigma_f^2\left[1 + \frac{1}{2\alpha} r^2(\x,\x^\prime) \right]^{-\alpha},
\end{equation}
and the ARD \emph{Mat\'ern 5/2} kernel \cite{snoek2012practical}, 
\begin{equation} \label{eq:matern}
\begin{split}
% k_\text{RQ}\left(\x,\x^\prime\right) = \sigma_f^2\left[1 + \frac{1}{2\alpha} r^2(\x,\x^\prime) \right]^{-\alpha} \qquad
k_\text{M52}\left(\x,\x^\prime\right) = & \sigma_f^2\left[1 + \sqrt{5 r^2(\x,\x^\prime)} + \frac{5}{3} r^2(\x,\x^\prime) \right] \exp\left\{ - \sqrt{5 r^2(\x,\x^\prime)} \right\},
%\text{with} \quad r^2(\x,\x^\prime) = & \sum_{d = 1}^{\nparams} \frac{1}{\ell_d^2} \left(x_d - x_d^\prime \right)^2,   \qquad \qquad \text{($\bm{\ell}$-scaled distance)}
\end{split}
\end{equation}
where $\sigma_f^2$ is the signal variance, and $\ell_1, \ldots, \ell_\nparams$ are the kernel length scales along each coordinate. 
Note that the RQ kernel tends to the SE kernel for $\alpha \rightarrow \infty$.

The Mat\'ern 5/2 kernel has become a more common choice for Bayesian \emph{global} optimization because it is only twice-differentiable \cite{snoek2012practical}, whereas the SE and RQ kernels are infinitely differentiable -- a stronger assumption of smoothness which may cause extrapolation issues. However, this is less of a problem for a local interpolating approximation (as in BADS) than it is for a global approach, and in fact we find the RQ kernel to work well empirically (see main text).

\paragraph{Composite periodic kernels}

We allow the user to specify one or more \emph{periodic} (equivalently, circular) coordinate dimensions $P \subseteq \{1, \ldots, \nparams \}$, which is a feature of some models in computational neuroscience (e.g., the preferred orientation of a neuron, as in the `neuronal selectivity' problem set \cite{goris2015origin} of the \ccn{} benchmark; see Section \ref{sec:ccn} in the main text). 
For a chosen base stationary covariance function  $k_0$ (e.g., RQ, SE, M$_{5/2}$), we define the composite ARD periodic kernel as
\begin{equation}
k_\text{PER}(\x, \x^\prime; k_0, P) = k_0\left(t(\x), t(\x^\prime) \right), \quad \text{with} \; \left\{ 
\begin{array}{rll} 
\left[ t(\x) \right]_{d} & =  x_d & \text{if} \; d \notin P \\
\left[ t(\x) \right]_d & =  \sin\left(\frac{\pi x_d}{L_d}\right) & \text{if} \; d \in P \\
\left[ t(\x) \right]_{d+|P|} & =  \cos\left(\frac{\pi x_d}{L_d}\right) & \text{if} \; d \in P 
\end{array}
\right.
\end{equation}
for $1\le d \le \nparams$, where $L_d$ is the period in the $d$-th coordinate dimension, and the length scale $\ell_d$ of $k_0$ is shared between $(d, d + |P|)$ pairs when $d \in P$. 
In BADS, the period is determined by the provided hard bounds as $L_d = \texttt{UB}_d - \texttt{LB}_d$ (where the hard bounds are required to be finite).

\subsection{Construction of the training set}

We construct the training set $\xx$ according to a simple \emph{subset-of-data} \cite{quinonero2007approximation} local GP approximation. 
Points are added to the training set sorted by their 
$\bm{\ell}$-scaled distance $r^2$ from the incumbent $\x_k$.
The training set contains a minimum of $n_\text{min} = 50$ points (if available in the cache of all points evaluated so far), and then up to $10 \times \nparams$ additional points with $r \le 3 \rho(\alpha)$, where $\rho(\alpha)$ is a \emph{radius function} that depends on the decay of the kernel. For a given stationary kernel of the form $k(\x, \x^\prime) = k(r^2(\x, \x^\prime))$, we define $\rho$ as the distance such that $k(2 \rho^2) \equiv 1/(\sigma_f^2 e)$. We have then
\begin{equation}
\rho_{SE} = 1, \qquad \rho_{M52} \approx 0.92, \quad \text{and} \quad \rho_{RQ}(\alpha) = \sqrt{\alpha (e^{1/\alpha}-1)},
\end{equation}
where for example $\rho_{RQ}(1) \approx 1.31$, and $\lim_{\alpha \rightarrow \infty} \rho_{RQ}(\alpha) = 1$.
 
\subsection{Treatment of hyperparameters}

We fit the GP hyperparameters by maximizing their posterior probability (MAP), $p(\vtheta | \xx, \y) \propto p(\vtheta, \xx, \y)$, which, thanks to the Gaussian likelihood, is available in closed form as \cite{rasmussen2006gaussian}
\begin{equation} \label{eq:logposterior}
\ln p(\y, \xx, \vtheta) = - \frac{1}{2}\ln |\K + \sigma^2 \I_n | - \frac{1}{2} \y^{\top} \left( \K + \sigma^2 \I_n \right)^{-1}\y + \ln p_\text{hyp}(\vtheta) + \text{const},
\end{equation}
where $\I_n$ is the identity matrix in dimension $n$ (the number of points in the training set), and $p_\text{hyp}(\vtheta)$ is the prior over hyperparameters, described in the following.

\paragraph{Hyperparameter prior}

We adopt an approximate \emph{empirical Bayes} approach by defining the prior based on the data in the training set, that is $p_\text{hyp} = p_\text{hyp}(\vtheta; \xx, \y)$. Empirical Bayes can be intended as a quick, heuristic approximation to a proper but more expensive hierarchical Bayesian approach.
We assume independent priors for each hyperparameter, with bounded (truncated) distributions.
Hyperparameter priors and hard bounds are reported in Table \ref{tab:hyperparams}. In BADS, we include an observation noise parameter $\sigma > 0$ also for deterministic objectives $f$, merely for the purpose of fitting the GP, since it has been shown to yield several advantages \cite{gramacy2012cases}. In particular, we assume a prior such that $\sigma$ decreases as a function of the poll size $\pollmesh$, as the optimization `zooms in' to smaller scales.  Another distinctive choice for BADS is that we set the mean for the GP mean equal to the 90-th percentile of the observed values in the current training set $\y$, which encourages the exploration to remain local.

\begin{table}[ht]
\resizebox{\textwidth}{!}{% % Adjust table font size
  \begin{tabular}{lll}
    \toprule
    %& \multicolumn{2}{c}{Algorithm}                   \\
    %\cmidrule{2-3}
    Hyperparameter     & Prior     & Bounds \\
    \midrule
    GP kernel \\
    \multicolumn{1}{r}{Length scales $\ell_d$} & $\ln \ell_d \sim \mathcal{N}_\text{T}\left(\frac{1}{2} (\ln r_{\text{max}} + \ln r_{\text{min}}), \frac{1}{4} (\ln r_{\text{max}} - \ln r_{\text{min}})^2 \right)$ & $[\Delta^\text{poll}_{\text{min}} , L_d]$      \\
        \multicolumn{1}{r}{Signal variability $\sigma_f$} & $\ln \sigma_f \sim \mathcal{N}_\text{T}\left(\ln \text{SD}(\y), 2^2 \right)$ & $[10^{-3} , 10^9]$      \\
        \multicolumn{1}{r}{RQ kernel shape $\alpha$} & $\ln \alpha \sim \mathcal{N}_\text{T}\left(1,1 \right)$ & $[-5, 5]$      \\
        \midrule
    GP observation noise $\sigma$ & $\ln \sigma \sim \mathcal{N}_\text{T}\left(\ln \sigmaest, 1 \right)$ & $[4 \cdot 10^{-4}, 150]$      \\
   \multicolumn{1}{r}{deterministic  $f$} & $\sigmaest = \sqrt{10^{-3} \pollmesh }$ &   \\
   \multicolumn{1}{r}{noisy  $f$} & $\sigmaest = $ 1 (or user-provided estimate) & \\
        \midrule
    GP mean $m$ & $m \sim \mathcal{N}\left(\text{Q}_{0.9}(\y),\frac{1}{5^2}(\text{Q}_{0.9}(\y) - \text{Q}_{0.5}(\y))^2 \right)$ & $(-\infty, \infty)$     
\\
    \bottomrule  
  \end{tabular}
  }
\vspace{1em}
  \centering
  \caption{{\bf GP hyperparameter priors.} Empirical Bayes priors and bounds for GP hyperparameters.  $\mathcal{N}\left(\mu, \sigma^2 \right)$ denotes the normal pdf with mean $\mu$ and variance $\sigma^2$, and $\mathcal{N}_\text{T}\left(\cdot,\cdot\right)$ the \emph{truncated} normal, defined within the bounds specified in the last column.  
   $r_{\text{max}}$ and $r_{\text{min}}$ are the maximum (resp., minimum) distance between any two points in the training set; $\Delta^\text{poll}_{\text{min}}$ is the minimum poll size (default $10^{-6}$); $L_d$ is the parameter range ($\texttt{UB}_d - \texttt{LB}_d$), for $1 \le d \le \nparams$; SD$(\cdot)$ denotes the standard deviation of a set of elements; $\pollmesh$ is the poll size parameter at the current iteration $k$; Q$_q(\cdot)$ denotes the $q$-th quantile of a set of elements (Q$_{0.5}$ is the median).} 
  \label{tab:hyperparams}
\end{table}

% \vspace{3em}
\paragraph{Hyperparameter optimization} We optimize Eq. \ref{eq:logposterior} with a gradient-based optimizer (see Section \ref{sec:implementation}), providing the analytical gradient to the algorithm. We start the optimization from the previous hyparameter values $\vtheta_{\text{prev}}$. If the optimization seems to be stuck in a high-noise mode, or we find an unusually low value for the GP mean $m$, we attempt a second fit starting from a draw from the prior averaged with $\vtheta_{\text{prev}}$. If the optimization fails due to numerical issues, we keep the previous value of the hyperparameters.
We refit the hyperparameters every $2 \nparams$ to $5 \nparams$ function evaluations; more often earlier in the optimization, and whenever the current GP is particularly inaccurate at predicting new points. We test accuracy on newly evaluated points via a Shapiro-Wilk normality test on the residuals \cite{royston1982extension}, $\zii = \left(\yii - \mu(\xii)\right)/\sqrt{\var(\xii) + \sigma^2}$ (assumed independent, in first approximation), and flag the approximation as inaccurate if $p < 10^{-6}$.

\subsection{Acquisition functions}

Besides the \emph{GP lower confidence bound} (LCB) metric \cite{srinivas2010gaussian} described in the main text (and default in BADS), 
we consider two other choices that are available in closed form using Eq. \ref{eq:gpmoments} for the GP predictive mean and variance. 

\paragraph{Probability of improvement (PI)}
This strategy maximizes the probability of improving over the current best minimum $\ybest$ \cite{kushner1964new}. 
For consistency with the main text, we define here the \emph{negative} PI,
\begin{equation}
a_\text{PI}\left(\x; \left\{\xx_n, \y_n \right\}, \vtheta\right) = -\Phi\left(\gamma(\x) \right), \qquad \gamma(\x) = \frac{\ybest - \xi - \mu\left(\x\right)}{s\left(\x\right)}
\end{equation}
where $\xi\ge 0$ is an optional trade-off parameter to promote exploration, and $\Phi\left( \cdot \right)$ is the cumulative distribution function of the standard normal. $a_\text{PI}$ is known to excessively favor exploitation over exploration, and it is difficult to find a correct setting for $\xi$ to offset this tendency  \cite{lizotte2008practical}.

\paragraph{Expected improvement (EI)} We then consider the popular predicted improvement criterion \cite{snoek2012practical,mockus1978application,jones1998efficient}. The expected improvement over the current best minimum $\ybest$ (with an offset $\xi \ge 0$) is defined as $\mathbb{E}\left[ \max\left\{\ybest - y, 0 \right\} \right]$. For consistency with the main text we consider the \emph{negative} EI, which can be computed in closed form as
\begin{equation} \label{eq:acqei}
a_\text{EI}\left(\x; \left\{\xx, \y \right\}, \vtheta\right) = - \sd\left(\x\right) \left[ \gamma(\x) \Phi\left(\gamma(\x) \right) + \mathcal{N}\left(\gamma(\x)\right) \right] 
\end{equation}
where $\mathcal{N}\left( \cdot \right)$ is the standard normal pdf.

\section{The BADS algorithm}
\label{sec:extrabads}

We report here extended details of the BADS algorithm, and how the various steps of the MADS framework are implemented (expanding on Sections \ref{sec:setup} and \ref{sec:bmads} of the main text). Main features of the algorithm are summarized in Table \ref{tab:badssum}. Refer also to Algorithm \ref{alg:bads} in the main text.

\begin{table}[ht]
  \centering
  \begin{tabular}{ll}
    \toprule
%    & \multicolumn{1}{c}{BADS}                   \\
%    \cmidrule{2-3}
    Feature     & Description (defaults)  \\
    \midrule
    Surrogate model     &  GP      \\
    Hyperparameter treatment  & optimization  \\
    GP training set size $n_\text{max}$ & 70 ($\nparams = 2$), 250 ($\nparams = 20$) (min 200 for noisy problems) \\
    \poll{} directions generation & LTMADS with GP rescaling  \\
    \search{} set generation  & Two-step ES algorithm with search matrix $\bm{\Sigma}$ \\
\search{} evals. ($n_\text{search}$) & $\max\{ \nparams, 3 + \lfloor \nparams/2 \rfloor \}$ \\
Aquisition function &  LCB \\     
Supported constraints & None, bound, and non-bound via a barrier function $c$ \\
Initial mesh size & $\meshz = 2^{-10}, \pollmesh = 1$ \\
Implementation & \texttt{bads} (MATLAB) \\
    \bottomrule  
  \end{tabular}
  \caption{{\bf Summary of features of BADS.}}
  \label{tab:badssum}
\end{table}

% \vspace{4em}
\subsection{Problem definition and initialization}
\label{subsec:probleminit}

BADS solves the optimization problem
\begin{equation}
\begin{split}
f_{\text{min}} = & \, \min_{x \in \X} f(\x) \qquad \text{with} \quad \X \subseteq \mathbb{R}^{\nparams}\\ 
\text{(\emph{optional})} \qquad c(\x) \le & \, 0
\end{split}
\end{equation}

where $\X$ is defined by pairs of hard bound constraints for each coordinate, $\texttt{LB}_d \le x_d \le \texttt{UB}_d$ for $1 \le d \le \nparams$, and we allow $\texttt{LB}_d \in \mathbb{R} \cup \{-\infty\}$ and similarly $\texttt{UB}_d \in \mathbb{R} \cup \{\infty\}$. We also consider optional non-bound constraints specified via a \emph{barrier} function $c: \X \rightarrow \mathbb{R}$ that returns constraint \emph{violations}. We only consider solutions such that $c$ is zero or less.

\paragraph{Algorithm input}

The algorithm takes as input a starting point $\x_0 \in \X$; vectors of \emph{hard} lower/upper bounds \texttt{LB}, \texttt{UB}; optional vectors of \emph{plausible} lower/upper bounds \texttt{PLB}, \texttt{PUB}; and an optional barrier function $c$.
We require that, if specified, $c(\x_0) \le 0$; and for each dimension $1\le d\le \nparams$, $\texttt{LB}_d \le (\x_0)_d \le \texttt{UB}_d$ and $\texttt{LB}_d \le \texttt{PLB}_d < \texttt{PUB}_d \le \texttt{UB}_d$. Plausible bounds identify a region in parameter space where most solutions are expected to lie, which in practice we usually think of as the region where starting points for the algorithm would be drawn from. 
Hard upper/lower bounds can be infinite, but plausible bounds need to be finite.
As an exception to the above bound ordering, the user can specify that a variable is \emph{fixed} by setting $(\x_0)_d = \texttt{LB}_d = \texttt{UB}_d = \texttt{PLB}_d = \texttt{PUB}_d$. Fixed variables become constants, and BADS runs on an optimization problem with reduced dimensionality.
The user can also specify circular or periodic dimensions (such as angles), which change the definition of the GP kernel as per Section \ref{sec:covariance}. 
The user can specify whether the objective $f$ is deterministic or noisy (stochastic), and in the latter case provide a coarse estimate of the noise (see Section \ref{sec:suppnoisy}).

\paragraph{Transformation of variables and constraints}

Problem variables whose hard bounds are strictly positive and $\texttt{UB}_d \ge 10 \cdot \texttt{LB}_d$ are automatically converted to log space for all internal calculations of the algorithm. All variables are also linearly rescaled to the standardized box $[-1,1]^\nparams$ such that the box bounds correspond to $\left[\texttt{PLB}, \texttt{PUB}\right]$ in the original space.
BADS converts points back to the original coordinate space when calling the target function $f$ or the barrier function $c$, and at the end of the optimization. BADS never violates constraints, by removing from the \poll{} and \search{} sets points that violate either bound or non-bound constraints ($c(\x) > 0$). During the \search{} stage, we project candidate points that violate a bound constraint to the closest mesh point within the bounds. We assume that $c(\cdot)$, if provided, is known and inexpensive to evaluate.

\paragraph{Objective scaling}

We assume that the scale of interest for differences in the objective (and the scale of other features, such as noise in the proximity of the solution) is of order $\sim1$, and that differences in the objective less than $10^{-3}$ are negligible.
For this reason, BADS is \emph{not} invariant to arbitrary rescalings of the objective $f$. 
This assumption does not limit the actual values taken by the objective across the optimization.
If the objective $f$ is the log likelihood of a dataset and model (e.g., summed over trials), these assumptions are generally satisfied. They would not be if, for example, one were to feed to BADS the \emph{average} log likelihood per trial, instead of the total (summed) log likelihood.
In cases in which $f$ has an unusual scale, we recommend to rescale the objective such that the magnitude of differences of interest becomes of order $\sim1$.

\paragraph{Initialization}

We initialize $\pollmeshz = 1$ and $\meshz =  2^{-10}$ (in standardized space).
The initial design comprises of the provided starting point $\x_0$ and $n_{\text{init}} = \nparams$ additional points chosen via a low-discrepancy Sobol quasirandom sequence \cite{bratley1988algorithm} in the standardized box, and forced to be on the mesh grid.
If the user does not specify whether $f$ is deterministic or stochastic, the algorithm assesses it by performing two consecutive evaluations at $\x_0$. For all practical purposes, a function is deemed noisy if the two evaluations at $\x_0$ differ more than $1.5 \cdot 10^{-11}$.\footnote{Since this simple test might fail, users are encouraged to actively specify whether the function is noisy.}

\subsection{\textsc{Search} stage}

In BADS we perform an aggressive \search{} stage in which, in practice, we keep evaluating candidate points  until we fail for $n_{\text{search}}$ consecutive steps to find a \emph{sufficient} improvement in function value, with $n_{\text{search}} = \max\{\nparams,\lfloor3+\nparams/2\rfloor\}$; and only then we switch to the \poll{} stage.
At any iteration $k$, we define an improvement \emph{sufficient} if $f_\text{prev} - f_\text{new} \ge (\pollmesh)^{3/2}$, where $\pollmesh$ is the poll size.

In each \search{} step we choose the final candidate point to evaluate, $\xsearch$, by performing a fast, approximate optimization of the chosen acquisition function in the neighborhood of the incumbent $\x_k$, using a two-step evolutionary heuristic inspired by CMA-ES \cite{hansen2003reducing}. This local search is governed by a \emph{search covariance matrix} $\bm{\Sigma}$, and works as follows.

\paragraph{Local search via two-step evolutionary strategy}

We draw a first generation of candidates $\bm{s}_{\text{\textsc{I}}}^{(i)} \sim \mathcal{N}(\x_k, (\pollmesh)^2 \bm{\Sigma})$ for $1 \le i \le n_\text{search}$, where we project each point onto the closest mesh point (see Section \ref{sec:mads} in the main text); $\bm{\Sigma}$ is a search covariance matrix with unit trace,\footnote{Unit trace (sum of diagonal entries) for $\bm{\Sigma}$ implies that a draw $\sim \mathcal{N}{(0, \bm{\Sigma})}$ has unit expected squared length.} and $n_\text{search} = 2^{11}$ by default. For each candidate point, we assign a number of offsprings inversely proportionally to the square root of its ranking according to $a(\bm{s}_{\text{\textsc{I}}}^{(i)})$, for a total of $n_\text{search}$ offsprings \cite{hansen2003reducing}. We then draw a second generation $\bm{s}_{\text{\textsc{II}}}^{(i)} \sim \mathcal{N}(\bm{s}_{\text{\textsc{I}}}^{(\pi_i)}, \lambda^2 (\pollmesh)^2 \bm{\Sigma})$ and project it onto the mesh grid, where $\pi_i$ is the index of the parent of the $i$-th candidate in the 2nd generation, and $0 < \lambda \le 1$ is a zooming factor (we choose $\lambda = 1/4$). Finally, we pick $\xsearch = \arg \min_i a(\bm{s}_{\text{\textsc{II}}}^{(i)})$. 
At each step, we remove candidate points that violate non-bound constraints ($c(\x) > 0$), and we project candidate points that fall outside hard bounds to the closest mesh point inside the bounds.

\paragraph{Hedge search} 

The search covariance matrix can be constructed in several ways. 
Across \search{} steps we use both a diagonal matrix $\bm{\Sigma}_{\bm{\ell}}$ with diagonal  $\left(\ell_1^2 / |\bm{\ell}|^2, \ldots, \ell_{\nparams}^2 / |\bm{\ell}|^2\right)$, and a matrix $\bm{\Sigma}_{\text{WCM}}$ proportional to the weighted covariance matrix of points in $\xx$ (each point weighted according to a function of its ranking in terms of objective values $y_i$, see \cite{hansen2003reducing}).
At each step, we compute the probability of choosing $\bm{\Sigma}_{s}$, with $s \in \{ \bm{\ell}, \text{WCM} \}$, according to a \emph{hedging} strategy taken from the \emph{Exp3} \textsc{Hedge} algorithm \cite{hoffman2011portfolio},
\begin{equation}
p_s = \frac{e^{\betahedge g_s}}{\sum_{s^{\prime}} e^{\betahedge g_{s^{\prime}}}} (1 - \gammahedge n_{\bm{\Sigma}}) + \gammahedge
\end{equation}
where $\betahedge = 1$, $\gammahedge = 0.125$, $n_{\bm{\Sigma}} = 2$ is the number of considered search matrices, and $g_s$ is a running estimate of the reward for option $s$. The running estimate is updated each \search{} step as
\begin{equation}
g_s^{\text{new}} = \alphahedge g_s^{\text{old}} + \frac{\Delta f_s}{p_s \pollmesh}
\end{equation}
where $\alphahedge = {0.1}^{1/(2 \nparams)}$ is a decay factor, and $\Delta f_s$ is the improvement in objective of the $s$-th strategy (0 if $s$ was not chosen in the current \search{} step).
This method  allows us to switch between searching along coordinate axes ($\bm{\Sigma}_{\bm{\ell}}$), and following an approximation of the local curvature around the incumbent ($\bm{\Sigma}_{\text{WCM}}$), according to their track record of cumulative improvement.

\subsection{\textsc{Poll} stage}
 
We perform the \poll{} stage only after a \search{} stage that did not produce a \emph{sufficient} improvement after $n_\text{search}$ steps.
We incorporate the GP approximation in the \poll{} in two ways: when constructing the set of polling directions $\D_k$, and when choosing the polling order. 

\paragraph{Set of polling directions}
At the beginning of the \poll{} stage, we generate a preliminary set of directions $\D^\prime_k$ according to the random LTMADS algorithm \cite{audet2006mesh}. We then transform it to a \emph{rescaled} set $\D_k$ based on the current GP kernel length scales: for $\bm{v}^\prime \in \D^\prime_k$, we define a rescaled vector $\bm{v}$ with $v_d \equiv v^\prime_d \cdot \omega_d$, for $1 \le d \le \nparams$, and $\omega_d \equiv \min \{ \max\{10^{-6}, \mesh, \ell_d / \text{GM}(\bm{\ell}) \}, \texttt{UB}_d - \texttt{LB}_d\}$, where $\text{GM}(\cdot)$ denotes the geometric mean, and we use $\texttt{PLB}_d$ (resp. $\texttt{PUB}_d$) whenever $\texttt{UB}_d$ (resp. $\texttt{LB}_d$) is unbounded. This construction of $\D_k$ deviates from the standard MADS framework. However, since the applied rescaling is bounded, we could redefine the mesh parameters and the set of polling directions to accomodate our procedure (as long as we appropriately discretize $\D_k$).
We remove from the poll set points that violate constraints, if present.

\paragraph{Polling order} Since the \poll{} is opportunistic, we evaluate points in the poll set starting from most promising, according to the ranking given by the chosen acquisition function \cite{gramacy2015mesh}.

\subsection{Update and termination}

If the \search{} stage was successful in finding a sufficient improvement, we skip the \poll{}, move the incumbent and start a new iteration, without changing the mesh size (note that mesh expansion under a success is not required in the MADS framework \cite{audet2006mesh}). If the \poll{} stage was executed, we verify if overall the iteration was successful or not, update the incumbent in case of success, and double (halven, in case of failure) the mesh size ($\tau = 2$). If the optimization has been \emph{stalling} (no sufficient improvement) for more than three iterations, we \emph{accelerate} the mesh contraction by temporarily switching to $\tau = 4$.

The optimization stops when one of these conditions is met:
\begin{itemize}
\item the poll size $\pollmesh$ goes below a threshold $\Delta^\text{poll}_{\text{min}}$ (default $10^{-6}$);
\item the maximum number of objective evaluations is reached (default $500 \times \nparams$);
\item the algorithm is \emph{stalling}, that is there has no \emph{sufficient} improvement of the objective $f$, for more than $4 + \lfloor \nparams/2 \rfloor$ iterations.
\end{itemize}

The algorithm returns the optimum $\x_{\text{end}}$ (transformed back to original coordinates) that has the lowest objective value $y_{\text{end}}$. For a noisy objective, we return instead the stored point with the lowest quantile $q_\beta$ across iterations, with $\beta = 0.999$; see Section \ref{sec:noisy} in the main text. We also return the function value at the optimum, $y_\text{end}$, or, for a noisy objective, our estimate thereof (see below, Section \ref{sec:suppnoisy}).
See the online documentation for more information about the returned outputs.

\subsection{Noisy objective}
\label{sec:suppnoisy}

For noisy objectives, we change the behavior and default parameters of the algorithm to offset measurement uncertainty and allow for an accurate local approximation of $f$. First, we:
\begin{itemize}
\item double the minimum number of points added to the GP training set, $n_\text{min} = 100$;
\item increase the total number of points (within radius $\rho$) to at least 200, regardless of $\nparams$;
\item increase the initial design set size to $n_\text{init} = 20$ points;
\item double the number of allowed stalled iterations before stopping.
\end{itemize}

\paragraph{Uncertainty handling} 
The main difference with a deterministic objective is that, 
due to observation noise, we cannot simply use the output values $y_i$ as ground truth in the \search{} and \poll{} stages. Instead, we adopt a \emph{plugin} approach \cite{picheny2013benchmark} and replace $y_i$ with the GP latent quantile function $q_\beta$ \cite{picheny2013quantile} (see Eq. 3 in the main text).
Moreover, we modify the MADS procedure  by keeping an \emph{incumbent set} 
$\{\x_i\}_{i=1}^k$, where $\x_i$ is the incumbent at the end of the $i$-th iteration. At the end of each \poll{} stage, we re-evaluate $q_\beta$ for all elements of the incumbent set, in light of the new points added to the cache which might change the GP prediction. We select as current (active) incumbent the point with lowest $q_\beta(\x_i)$. During optimization, we set $\beta = 0.5$ (mean prediction only), which promotes exploration. For the last iteration, we instead use a conservative $\beta_\text{end} = 0.999$ to select the optimum $\x_{\text{end}}$ returned by the algorithm in a robust manner. For a noisy objective, instead of the noisy measurement $y_\text{end}$, we return either our best GP prediction $\mu(\x_{\text{end}})$ and its uncertainty $\sd(\x_{\text{end}})$, or, more conservatively, an estimate of $\mathbb{E}[f(\x_{\text{end}})]$ and its standard error, obtained by averaging $N_\text{final}$ function evaluations at $\x_{\text{end}}$ (default $N_\text{final} = 10$). The latter approach is a safer option to obtain an unbiased value of $\mathbb{E}[f(\x_{\text{end}})]$, since the GP approximation may occasionally fail or have substantial bias.

\paragraph{Noise estimate} The user can optionally provide a noise estimate $\sigma_\text{est}$ which is used to set the mean of the hyperprior over the observation noise $\sigma$ (see Table \ref{tab:hyperparams}). We recommend to set $\sigma_\text{est}$ to the standard deviation of the noisy objective in the proximity of a good solution. If the problem has tunable precision (e.g., number of samples for log likelihoods evaluated via Monte Carlo), we recommend to set it, compatibly with computational cost, such that the standard deviation of noisy evaluations in the neighborhood of a good solution is of order 1.

\section{Benchmark}
\label{sec:benchmark}

We tested the performance of BADS on a large set of artificial and real problems and compared it with that of many optimization methods with implementation available in MATLAB (R2015b, R2017a).\footnote{MATLAB's \texttt{bayesopt} optimizer was tested on version R2017a, since it is not available for R2015b.} We include here details that expand on Section \ref{sec:design} of the main text.

\subsection{Algorithms}

\begin{table}[ht]
\resizebox{\textwidth}{!}{% % Adjust table font size
  \begin{tabular}{lllc|cc}
    \toprule
    Package     & Algorithm     & Source & Ref. & Noise & Global \\
    \midrule
\texttt{bads} & Bayesian Adaptive Direct Search & GitHub page \tablefootnote{\url{https://github.com/lacerbi/bads}} & This & \checkmark & \maybemark \\
\texttt{fminsearchbnd} & Nelder-Mead (\texttt{fminsearch}) w/ bounded domain & File Exchange\tablefootnote{\url{https://www.mathworks.com/matlabcentral/fileexchange/8277-fminsearchbnd--fminsearchcon}.} & \cite{lagarias1998convergence} & \xmark & \xmark
\\
\texttt{cmaes} & Covariance Matrix Adaptation Evolution Strategy
& Author's website\tablefootnote{\url{https://www.lri.fr/~hansen/cmaes_inmatlab.html}} & \cite{hansen2003reducing} & \xmark & \maybemark \\
--- (\texttt{active}) & CMA-ES with active covariance adaptation & --- & \cite{jastrebski2006improving} & \xmark & \maybemark \\
--- (\texttt{noise}) & CMA-ES with uncertainty handling & --- & \cite{hansen2009method} & \checkmark & \maybemark \\
\texttt{mcs} & Multilevel Coordinate Search & Author's website\tablefootnote{\url{https://www.mat.univie.ac.at/~neum/software/mcs/}} & \cite{huyer1999global} & \xmark & \checkmark \\
\texttt{snobfit} & Stable Noisy Optimization by Branch and FIT & Author's website\tablefootnote{\url{http://www.mat.univie.ac.at/~neum/software/snobfit/}} & \cite{huyer2008snobfit} & \checkmark & \checkmark \\
\texttt{global} & GLOBAL & Author's website\tablefootnote{\url{http://www.inf.u-szeged.hu/~csendes/index_en.html}} & \cite{csendes2008global} & \xmark & \checkmark \\
\texttt{randsearch} & Random search & GitHub page\tablefootnote{\url{https://github.com/lacerbi/neurobench/tree/master/matlab/algorithms}} & \cite{bergstra2012random} & \xmark & \checkmark \\
\midrule
\texttt{fmincon} & Interior point (\texttt{interior-point}, default) & Opt. Toolbox & \cite{waltz2006interior} & \xmark & \xmark \\
--- (\texttt{sqp}) & Sequential quadratic programming & --- & \cite{wright1999numerical} & \xmark & \xmark\\
--- (\texttt{active-set}) & Active-set & --- & \cite{gill1981practical} & \xmark & \xmark\\
\texttt{patternsearch} & Pattern search & Global Opt. Toolbox & \cite{kolda2003optimization} & \xmark & \xmark \\
\texttt{ga} & Genetic algorithms & Global Opt. Toolbox  & \cite{goldberg1989genetic} & \xmark & \maybemark \\
\texttt{particleswarm} & Particle swarm & Global Opt. Toolbox  & \cite{eberhart1995new} & \xmark & \maybemark \\
\texttt{simulannealbnd} & Simulated annealing w/ bounded domain & Global Opt. Toolbox  & \cite{kirkpatrick1983optimization} & \xmark & \maybemark \\
\texttt{bayesopt} & Vanilla Bayesian optimization & Stats. \& ML Toolbox  & \cite{snoek2012practical} & \checkmark & \checkmark \\
    \bottomrule  
  \end{tabular}
  }
\vspace{1em}
  \centering
  \caption{{\bf Tested algorithms.} \emph{Top}: Freely available algorithms. \emph{Bottom}: Algorithms in MATLAB's Optimization, Global Optimization, and Statistics and Machine Learning toolboxes. For all algorithms we note whether they explicitly deal with noisy objectives (\emph{noise} column), and whether they are local or global algorithms (\emph{global} column). Global methods (\checkmark) potentially search the full space, whereas local algorithms (\xmark) can only find a local optimum, and need a multi-start strategy. We denote with (\maybemark) \emph{semi-local} algorithms with intermediate behavior -- semi-local algorithms might be able to escape local minima, but still need a multi-start strategy.}
  \label{tab:algorithms}
\end{table}

The list of tested algorithms is reported in Table \ref{tab:algorithms}. For all methods, we used their default options unless stated otherwise. For BADS, CMA-ES, and \texttt{bayesopt}, we activated their \emph{uncertainty handling} option when dealing with noisy problems (for CMA-ES, see \cite{hansen2009method}).
For noisy problems of the \ccn{} set, within the \texttt{fmincon} family, we only tested the best representative method (\texttt{active-set}), since we found that these methods perform comparably to random search on noisy problems (see Fig \ref{fig:bbob09alt} right, and Fig \ref{fig:bbob09}, right panel, in the main text).
For the combinatorial game-playing problem subset in the \ccn{} test set, we used the settings of MCS provided by the authors as per the original study \cite{van2016people}.
We note that we developed algorithmic details and internal settings of BADS by testing it on the \textsc{cec14} test set for expensive optimization \cite{liang2013problem} and on other model-fitting problems which differ from the test problems presented in this benchmark.
For \texttt{bayesopt}, we allowed up to 300 training points for the GP, restarting the BO algorithm from scratch with a different initial design every 300 BO iterations (until the total budget of function evaluations was exhausted). The choice of 300 iterations already produced a large average algorithmic overhead of $\sim 8$ s per function evaluation. As acquisition function, we used the default EI-per-second \cite{snoek2012practical}, except for problems for which the computational cost is constant across all parameter space, for which we used the simple EI.
All algorithms in Table \ref{tab:algorithms} accept \emph{hard} bound constraints $\texttt{lb}$, $\texttt{ub}$, which were provided with the \bbob{} set and with the original studies in the \ccn{} set. For all studies in the \ccn{} set we also asked the original authors to provide \emph{plausible} lower/upper bounds $\texttt{plb}$, $\texttt{pub}$ for each parameter, which we would use for all problems in the set (if not available, we used the hard bounds instead). For all algorithms, plausible bounds were used to generate starting points. We also used plausible bounds (or their range) as inputs for algorithms that allow the user to provide additional information to guide the search, e.g. the length scale of the covariance matrix in CMA-ES, the initialization box for MCS, and  plausible bounds in BADS.

\subsection{Procedure}

For all problems and algorithms, for the purpose of our benchmark, we first transformed the problem variables according to the mapping described in `Transformation of variables and constraints' (Section \ref{subsec:probleminit}). In particular, this transformation maps the plausible region to the $[-1,1]^\nparams$ hypercube, and transforms to log space positive variables that span more than one order of magnitude. This way, all methods dealt with the same standardized domains.
Starting points during each optimization run were drawn uniformly randomly from inside the box of provided plausible bounds.

For deterministic problems, during each optimization run we kept track of the best (lowest) function value $\ybest^{t}$ found so far after $t$ function evaluations. We define the \emph{immediate regret} (or error) at time $t$ as $\ybest^{t} - \ymin$, where $\ymin$ is the true minimum or our best estimate thereof, and we use the error to judge whether the run is a success at step $t$ (error less than a given tolerance $\varepsilon$).
For problems in the \bbob{} set (both noiseless and noisy variants), we know the ground truth $\ymin$.
For problems in the \ccn{} set, we do not know $\ymin$, and we \emph{define} it as the minimum function value found across all optimization runs of all algorithms ($\approx 3.75 \cdot 10^5 \times \nparams$ function evaluations per noiseless problem), with the rationale that it would be hard to beat this computational effort. 
We report the \emph{effective} performance of an algorithm with non-negligible fractional overhead $o > 0$ by plotting at step $t \times o$ its performance at step $t$, which corresponds to a shift of the performance curve when $t$ is plotted in log scale (Fig \ref{fig:ccn17a} in the main text).\footnote{We did not apply this correction when plotting the results of vanilla BO (\texttt{bayesopt}), since the algorithm's  performance is already abysmal even without accounting for the substantial overhead.}

For noisy problems, we care about the true function value(s) at the point(s) returned by the algorithm, since, due to noise, it is possible for an algorithm to visit a neighborhood of the solution during the course of the optimization but then return another point.
For each noisy optimization run, we allowed each algorithm to return up to three solutions, obtained either from multiple sub-runs, or from additional outputs available from the algorithm, such as with MCS, or with population-based methods (CMA-ES, \texttt{ga}, and \texttt{particleswarm}). If more than three candidate solutions were available, we gave precedence to the main output of the algorithm, and then we took the two additional solutions with lowest observed function value.
We limited the number of candidates per optimization run to allow for a fair comparison between methods, since some methods only return one point and others potentially hundreds (e.g., \texttt{ga}) -- under the assumption that evaluating the true value of the log likelihood for a given candidate would be costly. 
For the combinatorial game-playing problem subset in the \ccn{} set, we increased the number of allowed solutions per run to 10 to match the strategy used in the original study \cite{van2016people}.
For noisy problems in the \ccn{} set, we estimated the log likelihood at each provided candidate solution via 200 function evaluations, and took the final estimate with lowest average.

For plotting, we determined ranking of the algorithms in the legend proportionally to the overall performance (area under the curve), across iterations (deterministic problems) or across error tolerances (noisy problems.)

\subsection{Alternative benchmark parameters}

\begin{figure}[htb]
  \includegraphics[width=\linewidth]{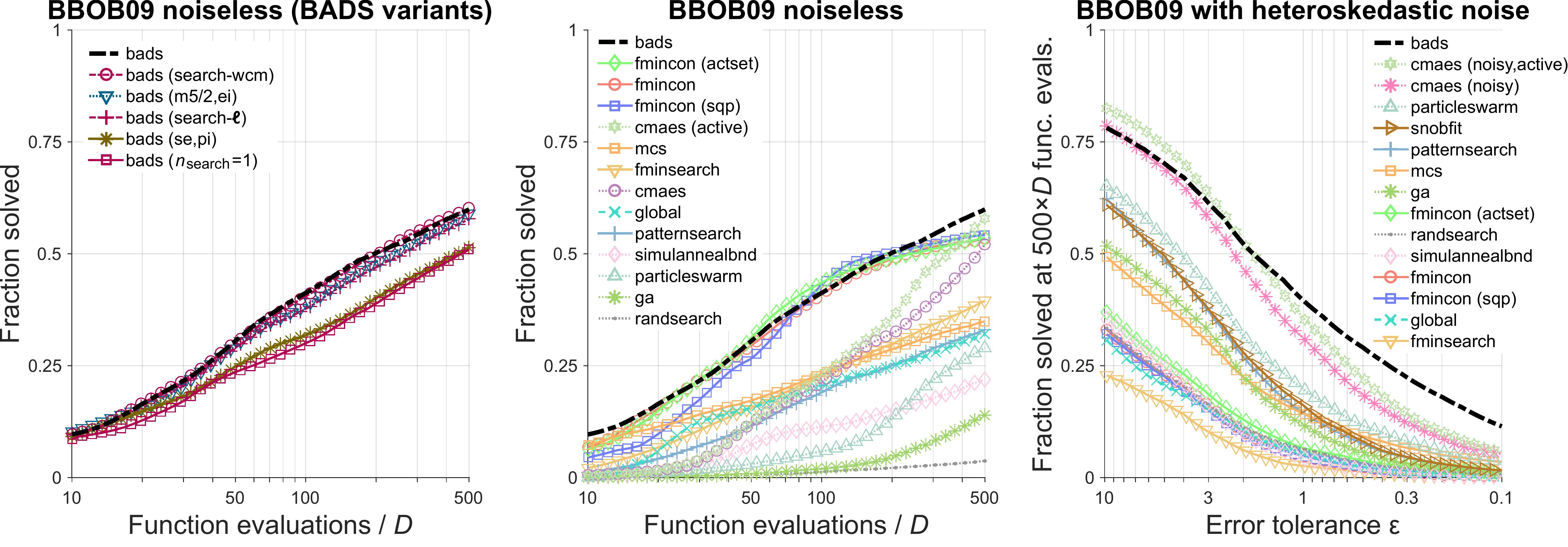}  
\vspace{-2em}
  \caption{{\bf Artificial test functions (\bbob{}).} Same as Fig \ref{fig:bbob09} in the main text, but with with alternative benchmark parameters (in bold). \emph{Left \& middle}: Noiseless functions. Fraction of successful runs ($\bm{\varepsilon \in [0.1,1]}$) vs. \# function evaluations per \# dimensions, for $\nparams \in \{3,6,10,15 \}$ (96 test functions); for different BADS configurations (\emph{left}) and all algorithms (\emph{middle}). \emph{Right}: Heteroskedastic noise. Fraction of successful runs at $\bm{500\times \nparams}$ objective evaluations vs. tolerance $\varepsilon$.}
  \label{fig:bbob09alt}
\end{figure}

In our benchmark, we made some relatively arbitrary choices to assess algorithmic performance, such as the range of tolerances $\varepsilon$ or the number of function evaluations. We show here that our findings are robust to variations in these parameters, by plotting results from the \bbob{} set with a few key changes (see Fig \ref{fig:bbob09} in the main text for comparison). First, we restrict the error tolerance range for deterministic functions to $\epsilon \in [0.1,1]$ instead of the wider range $\epsilon \in [0.01,10]$ used in the main text (Fig \ref{fig:bbob09alt} left and middle). This narrower range covers realistic practical requirements for model selection. Second, we reran the \bbob{} noisy benchmark, allowing $500 \times \nparams$ functions evaluation, as opposed to $200 \times \nparams$ in the main text (Fig \ref{fig:bbob09alt} right). Our main conclusions do not change, in that BADS performs on par with or better than other algorithms.
%which was chosen for tractability since noisy functions in the \ccn{} set are moderately expensive).

\section{Numerical implementation}
\label{sec:implementation}

BADS is currently freely available as a MATLAB toolbox, \texttt{bads} (a Python version is planned). 

The basic design of \texttt{bads} is simplicity and accessibility for the non-expert end user. First, we adopted an interface that resembles that of other common MATLAB optimizers, such as \texttt{fminsearch} or \texttt{fmincon}. Second, \texttt{bads} is \emph{plug-and-play}, with no requirements for installation of additional toolboxes or compiling \texttt{C/C++} code via \texttt{mex} files, which usually requires specific expertise.  Third, \texttt{bads} hides most of its complexity under the hood, providing the standard user with thoroughly tested default options that need no tweaking.

For the expert user or developer, \texttt{bads} has a modular design, such that \poll{} set generation, the \search{} oracle, acquisition functions (separately for \search{} and \poll{}), and initial design can be freely selected from a large list (under development), and new options are easy to add.

\paragraph{GP implementation}

We based our GP implementation in MATLAB on the GPML Toolbox \cite{rasmussen2010gaussian} (v3.6), modified for increased efficiency of some algorithmic steps, such as computation of gradients,\footnote{We note that version 4.0 of the GPML toolbox was released while BADS was in development. GPML v4.0 solved efficiency issues of previous versions, and might be supported in future versions of BADS.}, and we added specific functionalities.
We optimize the GP hyperparameters with \texttt{fmincon} in MATLAB (if the Optimization Toolbox is available), or otherwise via a the \texttt{minimize} function provided with the GPML package, modified to support bound constraints.

\end{document}